\documentclass[12pt]{article}

\usepackage{graphicx}
\usepackage{amssymb,amsmath}
\usepackage{amsthm}
\usepackage{paralist}
\usepackage{arydshln}
\usepackage[comma,authoryear]{natbib}
\RequirePackage{hypernat}

\RequirePackage{graphicx, rotating}
\RequirePackage{multirow}

\newtheorem{example}{Example}[section]
\newtheorem{theorem}{Theorem}[section]
\newtheorem{lemma}{Lemma}
\newtheorem{definition}{Definition}[section]

\newcommand{\bsb}{\boldsymbol}
\newcommand{\bsbX}{{\boldsymbol{X}}}
\newcommand{\bsbx}{{\boldsymbol{x}}}
\newcommand{\bsby}{{\boldsymbol{y}}}
\newcommand{\bsbb}{{\boldsymbol{\beta}}}
\newcommand{\bsbg}{{\boldsymbol{\gamma}}}

\newcommand{\bsbI}{{\boldsymbol{I}}}

\newcommand{\bsbSig}{{\boldsymbol{\Sigma}}}
\newcommand{\bsbxi}{{\boldsymbol{\xi}}}

\newcommand{\bsba}{{\boldsymbol{\alpha}}}

\newcommand{\bsbdelta}{{\boldsymbol{\delta}}}

\newcommand{\bsbmu}{{\boldsymbol{\mu}}}
\newcommand{\bsbW}{{\boldsymbol{W}}}

\newcommand{\bsbmI}{{\boldsymbol{\mathcal{I}}}}

\newcommand{\rd}{\,\mathrm{d}}

\begin{document}

%\begin{frontmatter}

\title{An Iterative Algorithm for Fitting  Nonconvex Penalized Generalized Linear Models with Grouped Predictors}
\date{}
\author{  Yiyuan She  \\
Department of Statistics\\ Florida State University,  FL 32306-4330, yshe@stat.fsu.edu}
\maketitle
%\author{Yiyuan She}
%\ead{yshe@stat.fsu.edu}
%\address{Department of Statistics, Florida State University, FL 32306-4330, United States}

%\vspace{-.5in}

\begin{abstract}
High-dimensional data pose challenges in statistical learning and modeling. Sometimes the predictors can be naturally grouped  where pursuing the between-group sparsity is desired.  Collinearity may occur  in real-world  high-dimensional applications where the popular $l_1$ technique suffers from both selection inconsistency and prediction inaccuracy. Moreover, the problems of interest often go beyond Gaussian models. To meet these challenges, nonconvex penalized generalized linear models with grouped predictors are investigated  and a simple-to-implement algorithm is proposed for  computation. A rigorous theoretical result guarantees its convergence  and provides tight preliminary scaling. This framework allows for grouped predictors and nonconvex penalties, including the discrete $l_0$  and the `$l_0+l_2$' type penalties. Penalty design and parameter tuning for nonconvex penalties are examined. Applications of   super-resolution spectrum estimation in signal processing and cancer classification with joint gene selection in bioinformatics show the performance improvement by  nonconvex penalized estimation.
\end{abstract}

%\vspace*{.3in}
%%
%%\begin{keyword}[class=AMS]
%%\kwd[Primary ]{62J12}
%%\kwd{68Q32}
%%\kwd[; secondary ]{62J07}
%%\end{keyword}
%

%\begin{keyword}
%%% keywords here, in the form: keyword \sep keyword
%
%%% MSC codes here, in the form: \MSC code \sep code
%%% or \MSC[2008] code \sep code (2000 is the default)
%
%%\noindent\textsc{Keywords}:
%{nonconvex penalties, group  lasso, generalized linear models, spectral analysis, gene selection}
%\end{keyword}
%\end{frontmatter}

%\vspace{-.3in}
\section{Introduction}
\label{secintro}
%\vspace{-.1in}
Penalized log-likelihood estimation is a useful technique in high-dimensional statistical modeling. Two basic and popular penalties are the $l_2$-penalty or ridge penalty, and the $l_1$-penalty or LASSO \citep{Tib}. Both are convex and are computationally feasible.  The ridge-penalty usually has the advantage of estimation  and prediction accuracy.
It is everywhere smooth and standard optimization methods such as  Newton-Raphson can be applied. By contrast,  the $l_1$-penalty is not differentiable at zero. This characteristic is however  useful and necessary in high-dimensional model selection, because exact zero components can be obtained in the LASSO estimate so that  a  number of nuisance features can be discarded. For the $l_1$ optimization algorithms in the Gaussian setup, refer to \citet{Efron}, \citet{Daub}, \citet{FHT}   among others.

On the other hand, the $l_1$-penalty cannot deal with \emph{collinearity}. Small coherence in the design, in form of the irrepresentable conditions \citep{Zhao}, RIP \citep{rip-lin}, sparse Riesz \citep{ZhangHuang} or others, is a \emph{must} for the $l_1$-type regularization to have good performance. 
Many real-world applications in signal processing and bioinformatics cannot fulfill this stringent requirement. For example, the super-resolution spectral estimation must apply an overcomplete dictionary at  fine enough frequency resolution and thus many sinusoidal atoms are highly correlated  (see Section \ref{sec:app}). %, and gene expression microarray data may contain tens of thousands of features, many of which are thought to be dependent and can be coexpressed.
When such collinearity occurs, (a) the prediction performance of the $l_1$-penalty is much worse than that of the $l_2$-penalty \citep{ZouHas}; (b) the sparsity recovery with the $l_1$ relaxation is inconsistent \citep{Zhao}.

%Its prediction performance is not good for correlated predictors~\citep{ZouHas}. Moreover, its variable selection is not guaranteed to  be consistent --- to achieve consistency, small coherence in the design matrix is a must,  see~\citet{Zhao},~\citet{DonohoSta},~\citet{Bunea},~\citet{ZhangHuang} among others.
%Yet many real-world applications in signal processing and bioinformatics   cannot fulfill this stringent requirement. For example,  the super-resolution spectral estimation must apply an overcomplete dictionary at a fine frequency resolution and thus many sinusoidal atoms are highly correlated, and gene expression microarray data may contain tens of thousands of features, many of which are thought to be dependent and can be coexpressed.

To see the necessity of applying nonconvex penalties, we remind that there are two objectives involved in the task of statistical learning and  modeling when one does not know the ground truth practically:
(\textbf{O1}) accurate prediction, and (\textbf{O2}) parsimonious model representation.
\textbf{O1}+ \textbf{O2} is consistent with Occam's razor principle. A good approach must reflect both concerns to produce a stable parsimonious   model  with  generalizability.

Seen from \textbf{O1},
a ridge penalty is desired  to account for noise and collinearity in the data. But it never encourages  sparsity. In the elastic net  which uses  a linear combination of the $l_1$ penalty and the $l_2$ penalty, the ridge part may  counteract the parsimony (\textbf{O2}) in the estimate  \citep{ZouHas}.  Yet the $l_1$-norm already provides the tightest convex relaxation of the $l_0$-norm. Therefore, to maintain accuracy \emph{and} promote sparsity, one must take into account nonconvex penalties such as those of type `$l_0+l_2$'.

This paper studies some computational problems in  statistical modeling  in the following setup:
\emph{
\begin{compactenum}
\item The data are high-dimensional, and correlated.
\item The predictors can be naturally grouped, where pursuing the between-group sparsity is desired.
\item A large family of penalties should be allowed for regularization, such as the $l_0$-penalty, $l_p$-penalties, and SCAD, in addition to the convex penalty family.
\item The methodology and analysis should go much beyond  Gaussian models to cover more applications such as classification.
\end{compactenum}
}

We briefly summarize some important (but absolutely not exhaustive) works  in the literature as follows.
In the Gaussian setup,
\citet{Daub} showed an iterative soft-thresholded procedure solves the $l_1$ penalized least-squares.  \citet{FHT} discovered a coordinate descent algorithm which can be viewed as a variant of the previous procedure.
Recently \citet{fhtglm} extended the algorithm to penalized generalized linear models (GLMs), by approximating the optimization problem at each iteration via  penalized weighted least-squares.
%and developed a popular {\tt R} package ({\tt glmnet}) to approximately solve the penalized generalized linear model (GLM) problem by penalized weighted least-squares at each step.
However, this approximation has no guarantee of convergence and may not provide a solution to the original problem.
These works focus on  convex penalties.

\citet{ZouLi} recently proposed the local linear approximation (LLA) for GLMs. An adaptive LASSO optimization is carried out at \emph{each} iteration step. The resulting algorithm has theoretical guarantee of convergence, but may not be efficient enough. %One-step LLA estimator is proposed, and is justified in the large $n$ setting.
Another popular approach is the DC programming  \citep{dcprog}, which  solves nonconvex penalized problems that can be represented as a difference of two convex functions \citep{FanLi,ZouLi,tong}. Similarly, a weighted LASSO problem is solved at each iteration. Neither of the techniques  directly  applies to discrete penalties, such as  $l_0$ and $l_0+l_2$, or group penalties.

To address the grouping concern,  \citet{Yuan}  proposed the group LASSO. An algorithm was developed under the assumption that the predictors within each group  are  \emph{orthogonal} to each other.
\citet{fhtnote} provided an algorithm for solving   convex group penalties in the Gaussian framework. How to address the nonconvex group penalties, e.g., the group $l_0+l_2$, for GLMs remains unsolved.

This paper provides a general framework for penalized log-likelihood optimization for any GLMs, to address all   points 1-4.
Our proposed algorithm significantly generalizes \citet{SheTISP} which was designed for Gaussian models only and could not attain discrete or group penalties.
Using a $q$-function trick, this framework allows for essentially any   penalties including the $l_0$, $l_p$, and SCAD penalties. The predictors can be grouped to pursue  the between-group sparsity. Moreover, the convergence analysis in this paper is less restrictive than \citet{SheTISP}. No condition is imposed on the penalty function. The proof is self-contained and the conclusion applies  to any thresholding rules (satisfying the  mild conditions given in Definition \ref{def:threshold}).

The rest of the paper is organized as follows.
Section 2 introduces the thresholding based algorithm with  rigorous theoretical convergence analysis and presents concrete penalty examples. %, and the obtained conditions are  more relaxed than those given  in the literature and lead to a decrease of the number of iterations in applications.  Concrete examples are given in Section 4, as well as some implementation details and variants including the relaxation form and the preliminary feature screening.
Section 3 discusses algorithm details and how to use numerical techniques and probabilistic screening for fast computation in high dimensions.
Section 4 investigates different choices of the penalty function by simulation studies, from which  a nonconvex hard-ridge penalty is advocated. Section 5 proposes a  selective cross-validation (SCV) scheme for parameter tuning. In Section 6, super-resolution spectrum reconstruction is studied and a real microarray data example is analyzed to illustrate the proposed methodology. % Further  discussions are given in Section 7.
Technical details are left to Appendix.
%%\vspace{-.1in}
%\section{Notation and Definitions}
%\label{sec:notation}
%%\vspace{-.1in}

\section{Solving  the Penalized Log-likelihood Estimation Problem}
\label{sec:algconv}
This paper assumes a \textbf{group GLM} setup that goes beyond Gaussianity.
Assume the observations $y_1,\cdots, y_n$ are independent and $y_i$ follows a distribution in the natural exponential family $f(y_i; \theta_i) = \exp (y_i\theta_i-b(\theta_i)+c(y_i))$, where $\theta_i$ is the natural parameter. Let $L_i=\log f(y_i, \theta_i)$, $L=\sum L_i$.
%Clearly, $L_i=y_i\theta_i-b(\theta_i)+c(y_i)$.
%, and thus ${\partial L_i}/{\partial \theta_i}=y_i - b'(\theta_i)$,
%${\partial^2 L_i}/{\partial \theta_i^2}=- b''(\theta_i)$. It is well known that $E(\partial L_i/\partial \theta_i)=0$ and $E(\partial L_i/\partial \theta_i)^2=-E(\partial^2 L_i/\partial \theta_i^2)$ hold in general for the exponential family. Therefore,
Then  $\mu_i\triangleq E(y_i)=b'(\theta_i)$.
%, $\mbox{var}(y_i)=b''(\theta_i)$.
Let $\bsbX=\left[\bsbx_1, \bsbx_2, \cdots, \bsbx_n\right]^T$ be the model matrix. The {canonical link} function, denoted by $g$, is applied. % such that $\bsbx_i^T\bsbb=g(\mu_i)=\theta_i$, and thus $g=(b')^{-1}$.
%For instance, when $y_i\sim \mbox{Bernoulli}(\pi_i)$, $f(y_i;\theta_i)=\exp \left\{ y_i\log\frac{\pi_i}{1-\pi_i}+\log(1-\pi_i)\right\}=\exp \left\{y_i \theta_i - \log(1+e^{\theta_i}) \right\}$, for which $\theta_i=\log \frac{\pi_i}{1-\pi_i}$, $\mu_i=\pi_i$, $b(t)=\log(1+e^{t})$,  and $g(t)=\log \frac{t}{1-t}$ (the logit link). In the Poisson case where $y_i\sim \mbox{Poi}(\omega_i)$, $f(y_i;\theta_i)=\frac{1}{y_i!} e^{-\omega_i} \omega_i^{y_i}=\exp(y_i\log \omega_i - \omega_i - \log y_i!)=\exp(y_i\theta_i-e^{\theta_i}+c(y_i))$ with $\theta_i=\log\omega_i$, $\mu_i=\omega_i$, $b(t)=e^t$, and $g(t)=\log t$ (the log link).
%Moreover, our studies can be trivially generalized  to the exponential dispersion family (see Section \ref{subsec:ext}) which covers the Gaussian model.
The Fisher information matrix %$[-\partial^2 L(\bsbb)/\partial \beta_h \beta_l]$
at $\bsbb$ is given  by $\bsbmI(\bsbb)=\bsbX^T\bsbW\bsbX$ with $\bsbW\triangleq \mbox{diag}\left\{b''(\bsbx_i^T\bsbb)\right\}$.
%%\vspace{-.1in}
%\section{Main Theorem of $\Theta$-estimators}
%%\vspace{-.1in}
%\label{sec:mainth}
%In this section, we consider penalized  generalized linear models (GLMs).
We assume the predictors are naturally\emph{grouped}, i.e., the design matrix is grouped into $K$ blocks: $\bsbX=[\bsbX_1, \cdots, \bsbX_K]\in {\mathbb R}^{n\times p}$,
so that in model selection one wants to keep or kill a group of predictors as a whole.
For a real example see the super-resolution spectral analysis in Section \ref{sec:app}. The predictor groups do not overlap but the group sizes can be different. When there are $p$ groups, each being  a singleton, the model reduces to the common `ungrouped' GLM.
The criterion of the group $P_k$-penalized log-likelihood  is defined by
%\vspace{-.1in}
\begin{eqnarray}
 F(\bsbb) \triangleq -L(\bsbb) + \sum_{k=1}^K P_{k}(\|\bsbb_k\|_2; {\lambda}_{k}), \label{oriprob-gen}
\end{eqnarray}
%\vspace{-.4in}
%
%\noindent
where $L=\sum_{i=1}^n L_i$ is the log-likelihood, $\bsbb_k$ are the coefficients associated with $\bsbX_k$, and $P_k$ are the penalty functions that can be discrete,  nonconvex, and nondifferentiable at zero.  % A parsimonious model is usually preferred in model interpretation especially when
The dimension $p$ may be much greater than the sample size $n$. There may exist a large number of nuisance features. % (in unit of groups).

Directly optimizing \eqref{oriprob-gen} can be tricky for a given penalty function. For example, the $l_0$-penalty $\frac{\lambda^2 }{  2} \|\bsbb\|_0=\frac{\lambda^2}{  2} |\{i: \beta_i \neq 0\}$ (where $| \cdot |$ is the set cardinality) used for building a parsimonious model is discrete and nonconvex. %To tackle the issue,
We turn to another class of estimators defined via an arbitrarily given thresholding rule to solve \eqref{oriprob-gen} for essentially any $P_k$.

\subsection{$\Theta$-estimators}
Somewhat interestingly, it is more convenient to tackle \eqref{oriprob-gen} from a thresholding viewpoint.
The main tool of this paper is the so-called \textit{$\Theta$-estimators}. First we define the thresholding rules rigorously as follows.
%Next we give a rigorous definition of the thresholding rule to be used as the main tool.
%\vspace{-.1in}
\begin{definition}[Threshold function]\label{def:threshold}
A threshold function is a real valued
function  $\Theta(t;\lambda)$ defined for $-\infty<t<\infty$
and $0\le\lambda<\infty$ such that
%\vspace{-.1in}
\begin{enumerate}%[\qquad\bf 1)]
\item $\Theta(-t;\lambda)= -\Theta(t;\lambda)$,
%\vspace{-.1in}
\item $\Theta(t;\lambda)\le \Theta(t';\lambda)$ for $t\le t'$, % we could also say $0\le t\le t'$
%\vspace{-.1in}
\item $\lim_{t\to\infty} \Theta(t;\lambda)=\infty$,\quad and
%\vspace{-.1in}
\item $0\le \Theta(t;\lambda)\le t$\ for\ $0\le t<\infty$.
\end{enumerate}
\end{definition}
%\vspace{-.1in}
In words, $\Theta(\cdot;\lambda)$  is an odd monotone unbounded shrinkage
rule for $t$, at any $\lambda$.  % $\Theta(0; \lambda)=0$ by definition.
A vector version of $\Theta$ (still denoted by $\Theta$) is defined componentwise if
either $t$ or $\lambda$ is replaced by a vector. %When both $t$ and $\lambda$ are vectors, we assume they have the same dimension.
Clearly,   $\Theta^{-1}(u;\lambda)\triangleq \sup\{t:\Theta(t;\lambda)\leq u\}, \forall u > 0$   must be monotonically increasing  and so its derivative is defined almost everywhere on $(0, \infty)$. For any $\Theta$, we introduce a finite positive constant ${\mathcal L}_{\Theta}$  such that $\rd\Theta^{-1}(u;\lambda)/\rd u$ is bounded below  almost everywhere  by $1- {\mathcal L}_{\Theta}$. For example, it is easy to show ${\mathcal L}_{\Theta}$ can be $0$ and $1$ for soft-thresholding and hard-thresholding respectively.

A \emph{multivariate} version of $\Theta$, denoted by $\vec\Theta$, is defined for any vector $\bsba\in \mathbb R^p$:
\begin{eqnarray}
\vec\Theta(\bsba;\lambda)=\bsba^{\circ} \Theta(\|\bsba\|_2;\lambda),   \label{multithetadef}
\end{eqnarray}
%\vspace{-.5in}
%
%\noindent
where
%\vspace{-.1in}
$
\bsba^{\circ}=\begin{cases}\frac{\bsba}{\|\bsba\|_2}, & \mbox{ if }  \bsba\neq \bsb{0}\\ \bsb{0}, & \mbox{ if }  \bsba= \bsb{0} \end{cases}
$.
%
%%\vspace{-.1in}
%\begin{definition}[Multivariate threshold function]\label{def:thresholdvec}
%Given any  threshold function $\Theta(\cdot; \lambda)$ with $\lambda$ as the parameter, we define its multivariate version $\vec\Theta$ by
%%\vspace{-.2in}
%\begin{eqnarray}
%\vec\Theta(\bsba;\lambda)=\bsba^{\circ} \Theta(\|\bsba\|_2;\lambda), \quad \forall \bsba\in {\mathbb R}^n \label{multithetadef}
%\end{eqnarray}
%%\vspace{-.5in}
%%
%%\noindent
%where
%%\vspace{-.1in}
%$$
%\bsba^{\circ}=\begin{cases}\frac{\bsba}{\|\bsba\|_2}, & \mbox{ if }  \bsba\neq \bsb{0}\\ \bsb{0}, & \mbox{ if }  \bsba= \bsb{0}.\end{cases}
%$$
%\end{definition}
%%\vspace{-.1in}
%
Obviously, $\vec\Theta$ is still a shrinkage rule because $\|\vec\Theta(\bsba;\lambda)\|_2= \Theta(\|\bsba\|_2;\lambda) \leq \|\bsba\|_2$. \\
%\vspace{-.2in}
%$$
%s'\geq -L_{\Theta} \quad a.e.
%\vspace{-.2in}
%$$
%Seen from \eqref{defofthetainv}, $s'\geq -1$ almost everywhere and so
%Clearly, a finite $L_{\Theta}$ exists.

%Thresholding rules  are associated with penalty functions.
%There may exist multiple (or even infinitely many) penalties that correspond to the same threshold function.
%The following \emph{three-step construction} finds the penalty with the smallest curvature~\citep{antrev,SheTISP}:
%%\vspace{-.25in}
%\begin{equation}\label{defofthetainv}
%\begin{split}
%\Theta^{-1}(u;\lambda)&=\sup\{t:\Theta(t;\lambda)\leq u\},  %\mbox{ and } \Theta^{-1}(-u;\lambda)=-\Theta^{-1}(u;\lambda),
%\\
%s(u;\lambda) &=  \Theta^{-1}(u;\lambda)-u,\quad\text{and} \\
%P(\theta;\lambda)&=P_\Theta(\theta;\lambda)=\int_0^{|\theta|} s(u;\lambda)\rd u,
%\end{split}
%\end{equation}
%%\vspace{-.35in}
%%
%%\noindent
%where $u\ge 0$ holds throughout~\eqref{defofthetainv}.
%The constructed penalty $P_\Theta$ is nonnegative and is continuous in $\theta$.

Now we define \emph{group $\Theta$-estimators}. Given any threshold functions  $\Theta_1,\ldots,\Theta_K$,  the induced group $\Theta$-estimator satisfies the following nonlinear equation
%\vspace{-.3in}
\begin{eqnarray}
\bsbb_k = \vec\Theta_k(\bsbb_k+\bsbX_k^T\bsby-\bsbX_k^T \bsbmu(\bsbb); {\lambda}_k), \quad 1\leq k \leq K,
\label{thetaestglm-gen}
\end{eqnarray}
%\vspace{-.45in}
%
%\noindent
where $\mu_i=g^{-1}(\bsbx_i^T\bsbb)$ with $g$ as the canonical link function. 
To avoid the influence of the ambiguity in defining some threshold functions (e.g., hard-thresholding), we always assume the quantity to be thresholded does not correspond to any discontinuity of $\vec\Theta_k$. This assumption is mild because a practical thresholding rule usually has at most finitely many discontinuity points and such discontinuities rarely occur in any real application.

As will be shown later, there is a universal connection between the penalized estimators and the group $\Theta$-estimators, but the latter are much easier to compute:
%We introduce the GLM version of the group TISP which generalizes~\citet{SheTISP}:
at each iteration step $j$, the new $\bsbb^{(j+1)}$ can be updated through the multivariate thresholding
%\vspace{-.15in}
\begin{align}
 {Group\mbox{-}TISP\mbox{:}} \ \ \bsbb_k^{(j+1)} = \vec\Theta_k(\bsbb_k^{(j)}+\bsbX_k^T\bsby-\bsbX_k^T \bsbmu(\bsbb^{(j)}); {\lambda}_k), 1\leq k \leq K,
\label{tisp-gen}
\end{align}
%\vspace{-.45in}
%
%\noindent
provided that the norm of the global design $\bsbX$ is not  large (as will be explained in Theorem \ref{conv-gen}).
This suggests the need of scaling the data beforehand (which does not affect the sparsity of $\bsbb$).
We refer to \eqref{tisp-gen}  as group thresholding-based iterative selection procedure (\emph{Group TISP}). It  generalizes the work by \citet{SheTISP} in the Gaussian nongrouped setup.
%Note that we do not handle GLMs by approximate weighted least squares as in~\citet{fhtglm} which may not provide a solution to the original problem and may diverge in our experience.
Next, we show \eqref{tisp-gen} converges properly to a group $\Theta$-estimate under some appropriate conditions, which in turn solves the penalized log-likelihood problem  \eqref{oriprob-gen} in a general sense.
%Recall that $\bsbmI(\bsbb)$ stands for the Fisher information matrix at $\bsbb$ and $L_\Theta$ is a constant associated with a threshold function $\Theta$ (see Section \ref{sec:notation}).
%The following theorem answers the question between penalized log-likelihoods, $\Theta$-estimators, and TISP.

\begin{theorem}
\label{conv-gen}
Let $\Theta_{k}$ ($1\leq k \leq K$) be arbitrarily given thresholding rules and  $\bsbb^{(0)}$ be any $p$-dimensional vector. Denote by  $\bsbb^{(j)}, j=1, 2, \cdots$, the group TISP  iterates  %induced by $(\vec{\Theta}_{1},\vec{\Theta}_{2}, \cdots, \vec{\Theta}_{K})$
defined via \eqref{tisp-gen}.
%Denote by $A$ the set of $\{t\bsbb^{(j)}+(1-t)\bsbb^{(j+1)}: t \in (0, 1), j=1, 2, \cdots\}$, and
Define $\rho=\sup_{\bsbxi\in A} \|\bsbmI(\bsbxi)\|_2$ where $A= \{\vartheta \bsbb^{(j)}+(1-\vartheta)\bsbb^{(j+1)}:\vartheta  \in (0, 1), j=1, 2, \cdots\}$.
%Define $L_{\Theta_1,\cdots, \Theta_K}=\max_{1\leq k \leq K} L_{\Theta_{k}}$. Then,
If
\begin{align}
\rho \leq \max(1, {2-\max_{1\leq k \leq K} {\mathcal L}_{\Theta_k}}),
\end{align}
then for any penalty functions $P_{k}$  satisfying
%\vspace{-.2in}
$$
P_{k}(\theta;\lambda_k)-P_{k}(0;\lambda_k)=
\int_0^{|\theta|} (\sup\{s:\Theta_k(s;\lambda_k)\leq u\} - u) \rd u
%P_{\Theta_k}(\theta; \lambda_k)
+ q_{k}(\theta; \lambda_k),
%\vspace{-.2in}
$$
%where $P_{\Theta_k}$ is obtained via \eqref{defofthetainv},
with  $q_{k}(\theta,\lambda_k)$   nonnegative  and  $q_{k}(\Theta_{k}(t;\lambda_k);\lambda_k)=0, \forall t\in \mathbb R$,
the value of the corresponding  objective function  $F$ in \eqref{oriprob-gen} decreases at each iteration
%\setlength{\arraycolsep}{5pt}
%\vspace{-.15in}
\begin{align}
F(\bsbb^{(j)})- F(\bsbb^{(j+1)})\geq C \|\bsbb^{(j)}-\bsbb^{(j+1)}\|_2^2, \quad j=1,2,\cdots
\label{asympreg-gen}
\end{align}
%\vspace{-.5in}
%
%\noindent
where $C= \max(1, 2-\max_k {\mathcal L}_{\Theta_k})- \rho$.
If, further, $\rho < \max(1, {2-\max_k {\mathcal L}_{\Theta_k}})$,  any limit point of  $\bsbb^{(j)}$ must be a fixed point of \eqref{thetaestglm-gen}, or a group $\Theta$-estimate.
\end{theorem}

See  \ref{app} for its proof.
The theorem allows for $p>n$ and applies to \emph{any} threshold functions, even if they are not nonexpansive.
This covers essentially any penalties of practical interest, as will be shown below.

%%\vspace{-.1in}
%\section{TISP for Generalized Linear Models}
%\label{sectispglm}
%%\vspace{-.1in}
\subsection{Concrete examples}
%\vspace{-.05in}
The theorem  indicates no matter how the predictors are grouped, for an arbitrarily given model matrix, performing a simple preliminary scaling $\bsbX/k_0$   always guarantees the convergence of the algorithm of \eqref{tisp-gen}, provided $k_0$ is appropriately large. For a specific GLM, the choice of $k_0$ can be made regardless of $\Theta$, $\lambda$, and $K$.
%\vspace{-.1in}
\begin{example}[Gaussian GLM] \label{ex:gauss}
If $y_i$ are Gaussian,  $\bsbmu(\bsbb) = \bsbX\bsbb$ in \eqref{tisp-gen} and  $\bsbmI=\bsbSig=\bsbX^T\bsbX$. Therefore,  $k_0\geq \| \bsbX\|_2$ suffices regardless of the specific thresholding rules. This covers \citet{SheTISP} where  the predictors are ungrouped ($K=p$) and all $\Theta_k$'s are identical.
\vspace{-.1in}
\end{example}

\begin{example}[Binomial GLM]  \label{ex:bin}
If $y_i\sim \mbox{Bernoulli}(\pi_i)$ as in classification problems, we can write $\bsbmu(\bsbb)$ as $1/(1+\exp(-\bsbX\bsbb))$ with the  operations being elementwise except for the matrix-vector multiplication of $\bsbX \bsbb$. Now the proposed algorithm reduces to
%\vspace{-.15in}
\begin{eqnarray}
\bsbb_k^{(j+1)} = \vec\Theta_k\left(\bsbb_k^{(j)} + \bsbX_k^T\bsby - \bsbX_k^T \left[\frac{1}{1+\exp(-\bsbX\bsbb^{(j)})}\right]_{n\times p}; \lambda_k\right), 1\leq k \leq K.
\end{eqnarray}
%\vspace{-.45in}
%
%\noindent
For \emph{ungrouped} predictors ($K=p$) and identical $\Theta_k$'s, the iteration can be simplified to
%\vspace{-.15in}
\begin{eqnarray}
\bsbb^{(j+1)} = \Theta\left(\bsbb^{(j)} + \bsbX^T\bsby - \bsbX^T \left[\frac{1}{1+\exp(-\bsbX\bsbb^{(j)})}\right]_{n\times p}; \lambda \right). \label{logit-tisp}
\end{eqnarray}
%\vspace{-.4in}
%
%\noindent
In either case, since $w_i=b''(\bsbx_i^T\bsbb)=\pi_i(1-\pi_i)\leq1/4$, a somewhat crude but general choice is $k_0\geq \|\bsbX\|_2/2$, regardless of $\Theta$.  The procedure based on \eqref{logit-tisp} is different than the algorithm in \citet{fhtglm} that approximates the original penalized logistic regression problem by penalized weighted least-squares at each iteration. Our algorithm has theoretical guarantee of convergence.  %Theorem \ref{conv-gen} also guarantees the convergence of \eqref{logit-tisp} to a stationary point of the original optimization problem.
%\vspace{-.1in}
\end{example}

\begin{figure}[h!]
\begin{center}
\includegraphics[width=5.5in]{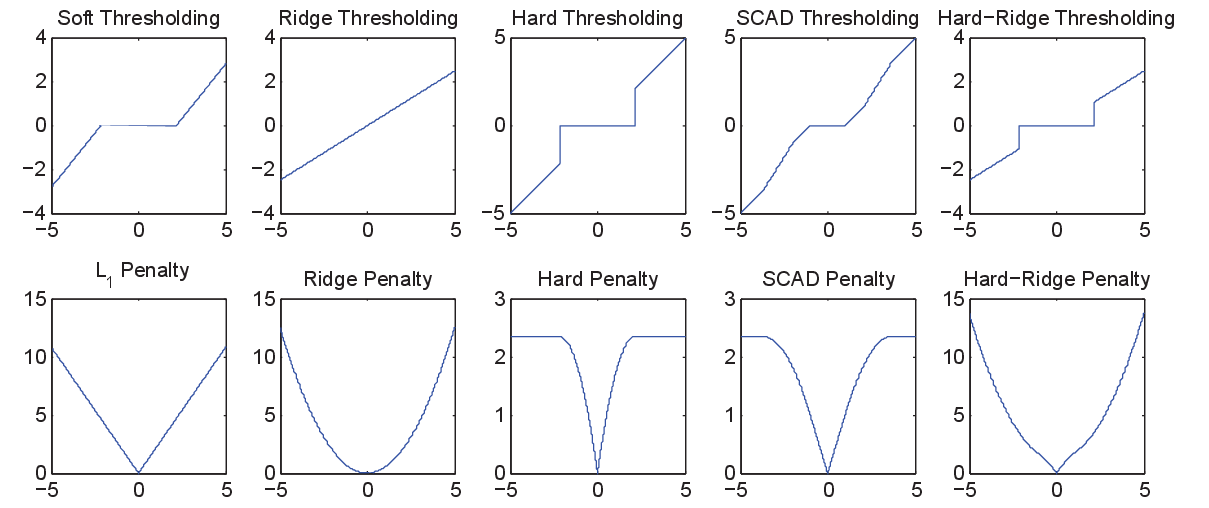}
\end{center}
\caption[Some penalties]{\small{Some examples of the thresholding rules and their corresponding penalties. Left to right: Soft, Ridge, Hard, SCAD, and Hard-ridge. }}
\label{figthpens}
\end{figure}

On the other hand,  the experience indicates that if the algorithm converges, smaller values of $k_0$  lead  to faster convergence. It is a meaningful question in  computation to find the \emph{least} possible $k_0$  in concrete applications. Theorem \ref{conv-gen} provides useful guidance in this regard: the  $\rho$-bound based on ${\mathcal L}_{\Theta}$ seems to be  tight enough in implementation for various $\Theta$. In the following, we give some examples of $\Theta$ and $P$ to show the power of the proposed algorithm for solving penalized likelihood estimation. See Figure \ref{figthpens} for an illustration.
The function $q$ in the theorem is often 0, but we use nontrivial $q$'s in Example \ref{ex:hard} and Example \ref{ex:hybrid} to attain the discrete $l_0$ penalty and the $l_0+l_2$ penalty.

%In the Poisson case, $w_i=\exp(\bsbx_i^T\bsbb)$, $i=1,\cdots, n$, for which we do not have a universal bound. However, in practice, it is usually easy to find a large scaling constant  to guarantee the convergence in concrete problems. For example, the unpenalized MLE estimate may act as a good reference.

%There are rich examples of $\Theta$ to be used. %, such as the soft-thresholding, hard-thresholding, SCAD-thresholding, and the hybrid hard-ridge-thresholding.
%See Figure \ref{figthpens} for an illustration.
%The function $q$ in the theorem is often 0, but we use nontrivial $q$ in Example \ref{ex:hard} and Example \ref{ex:hybrid} to derive  multiple penalties (including the discrete $l_0$-penalty) all resulting in the same $\Theta$-estimator. %Therefore, a thresholding launching point might be appropriate in looking for the best sparse estimator, as advocated by~\citet{SheTISP}.
%
%
%\vspace{-.1in}

\begin{example}[$L_1$] \label{ex:soft}
When $\Theta$ is the soft-thresholding -- $\Theta_S(t;\lambda)=\mbox{sgn}(t) (|t|-\lambda) 1_{|t|\geq \lambda}$, the associated penalty is $P(\theta; \lambda)=\lambda  |\theta |$. Since we can set ${\mathcal L}_{\Theta}=0$, the scaling constant can be relaxed to $k_0=\|\bsbX\|_2/\sqrt{2}$ in regression and  $k_0=\|\bsbX\|_2/(2\sqrt{2})$ in classification.
For grouped predictors, the algorithm of \eqref{tisp-gen} solves Problem \eqref{oriprob-gen} with the group $l_1$-penalty $\sum_k \lambda_k \|\bsbb_k\|_2$ for any GLM, the scaling constant being the same.
%This extends the grouped LASSO~\citep{Yuan} to  GLMs.
In comparison to \citet{Yuan}, we do \emph{not} have to make the simplistic  assumption that the predictors must be orthogonal to each other within each group.
%$\bsbX_k$ have been (column) \emph{orthogonalized} within each group,
%It is interesting to note that the asynchronous component-by-component updating of \eqref{tisp}  leads  to the coordinate decent algorithm~\citep{FHT} in the Gaussian setup. The corresponding pathwise algorithm is fast for solving the lasso problem when $p>n$.
\vspace{-.1in}
\end{example}

\begin{example}[Elastic net] \label{ex:enet}
Define  $\Theta(t;\lambda_1, \lambda_2)\triangleq \Theta_S(\frac{t}{1+\lambda_2}; \frac{\lambda_1}{1+\lambda_2})$, where $\Theta_S$ is the soft-thresholding. Then the elastic net \citep{ZouHas}  problem is solved, where  $P(\theta; \lambda_1, \lambda_2)=\lambda_1 |\theta| + \lambda_2 \theta^2/2$. %Though inducing no sparsity, the $l_2$-penalty often leads to better accuracy in estimation and prediction.
\vspace{-.1in}
\end{example}
\begin{example}[$L_0$] \label{ex:hard}
Let $\Theta$ be the hard-thresholding $t 1_{|t|\geq \lambda}$. Then ${\mathcal L}_{\Theta}=1$.  According to the theorem, letting $q\equiv 0$, our algorithm solves for the `hard penalty'
%\vspace{-.2in}
\begin{eqnarray}
P_{H}(\theta; \lambda)= \begin{cases} -\theta^2/2+\lambda |\theta|, & \mbox{ if } |\theta|<\lambda\\ \lambda^2/2,  &\mbox{ if } |\theta|\geq \lambda. \end{cases} \label{hardpen}
\end{eqnarray}
%\vspace{-.35in}
%
%\noindent
Interestingly, setting
%\vspace{-.15in}
$$
q(\theta;\lambda)=\begin{cases} \frac{(\lambda-|\theta|)^2}{2} , & \mbox{ if } 0 < |\theta| < \lambda\\
0,  & \mbox{ if }  \theta=0 \mbox{ or } |\theta| \geq \lambda, \end{cases}
$$
%\vspace{-.35in}
%
%\noindent
we obtain the discrete $l_0$-penalty $P(\theta; \lambda)=\frac{\lambda^2}{2} 1_{\theta\neq 0} $.
%In fact, our theorem implies there exist infinitely many penalties all mimicking the $l_0$-penalty and resulting in the same $\Theta$-estimate.
Similarly, we can justify that the continuous penalty $P(\theta;\lambda)=\alpha P_H(\theta; \lambda/\sqrt\alpha)$ mimics the $l_0$-penalty and results in the same $\Theta$-estimate, for any $\alpha \geq  1$.
For grouped predictors, our algorithm provides a solution to  the group $l_0$-penalty $\sum_{k=1}^K \frac{\lambda_k^2}{2} 1_{\|\bsbb_k\| \neq 0}$ which can attain more  between-group sparsity than the group LASSO.
\vspace{-.1in}
\end{example}
\begin{example}[Firm \& SCAD] \label{ex:scad}
The firm shrinkage~\citep{firm} is defined by
%\vspace{-.15in}
\begin{eqnarray}
\Theta(t; \lambda, \alpha)=\begin{cases}0, & \mbox{ if } |t|<\alpha \lambda\\ \frac{t-\alpha\lambda\mbox{sgn}(t)}{1-\alpha}, & \mbox{ if } \alpha \lambda \leq   |t| < \lambda\\
t, & \mbox{ if }     |t|  \geq \lambda,
\end{cases}
\label{firmth}
\end{eqnarray}
%\vspace{-.35in}
%
%\noindent
where $0\leq \alpha \leq 1$. % (For $\alpha\geq 1$, see Example 3.)
The penalty function  is then  $\alpha  P_H(t; \lambda)$. An equivalent form of this penalty is used in MCP~\citep{mplus}.
%Suppose  $\Theta$ is the SCAD-thresholding. Then we get the SCAD-penalty whose derivative is defined by
%\vspace{-.15in}
A related thresholding is the SCAD-thresholding~\citep{FanLi} and the SCAD-penalized GLMs with grouped predictors can be solved by \eqref{tisp-gen}.
%\begin{eqnarray}
%{P'}(\theta;\lambda) = \begin{cases} \lambda, & \mbox{ if } \theta\leq\lambda\\ (a\lambda-\theta)/(a-1), & \mbox{ if } \lambda<\theta\leq a\lambda\\ 0, & \mbox{ if } \theta > a \lambda\end{cases}
%\end{eqnarray}
%%\vspace{-.35in}
%%
%%\noindent
%for $\theta>0$ and $a>2$.
%$L_{\Theta}$ can be $1/(a-1)$. By default, $a=3.7$~\citep{FanLi}.

\vspace{-.1in}
\end{example}

\begin{example}[$L_\mathrm p$] \label{ex:lp}
We focus on $0<\mathrm p<1$. Assuming $\lambda\geq 0$, define a function $$g(\theta; \lambda) = \theta + \lambda \mathrm p \theta^{\mathrm p-1}$$ for any $\theta \in [0, +\infty)$. It is easy to verify that (i) $g$ attains its minimum $\tau(\lambda)=\lambda^{1/(2-\mathrm p)} (2-\mathrm p) [\mathrm p/(1-\mathrm p)^{1-\mathrm p}]^{1/(2-\mathrm p)}$ at $\theta_o=\lambda^{1/(2-\mathrm p)} [\mathrm p(1-\mathrm p)]^{1/(2-\mathrm p)}$; (ii) $g(\theta)$ is strictly increasing on $[\theta_o, +\infty)$; (iii) $g(\theta)\rightarrow +\infty$ as $\theta \rightarrow +\infty$.  Therefore, given any $t>\tau(\lambda)$, the  equation $g(\theta) = t$ has one and only one root in $[\theta_o,+\infty)$ (or $[\theta_o,t)$, as a matter of fact), which can be found numerically.
Given $\mathrm p\in(0, 1)$, introduce the following function
\vspace{-.1in}
\begin{align}
\Theta_{l_\mathrm p}(t; \lambda) = \begin{cases}0,& \mbox{ if } |t| \leq \tau(\lambda)\\
    \mbox{sgn}(t)\max\{\theta: g(\theta) = |t|\}, & \mbox{ if } |t| > \tau(\lambda). \end{cases}
\end{align}
Based on the properties of $g$, it is not difficult to show that $\Theta_{l_\mathrm p}(\cdot;\lambda)$ is indeed a threshold function, i.e., an odd monotone unbounded shrinkage rule. From the theorem, $\Theta_{l_\mathrm p}$ can handle $P_{\Theta_{l_\mathrm p}}(\theta; \lambda) = \lambda |\theta|^\mathrm p$. %Hence the $\Theta_{l_p}^\sigma$ gives the exact solution of the Schatten $p$-norm (with no root)  penalized matrix approximation.
\vspace{-.05in}
\end{example}

\begin{example}[Hard-ridge ($L_0+L_2$)] \label{ex:hybrid}
The hybrid hard-ridge-thresholding is defined based on the hard-thresholding and the ridge-thresholding  \citep{SheTISP}
%\vspace{-.15in}
\begin{eqnarray}
\Theta(t;\lambda,\eta)=\begin{cases} 0, & \mbox{ if } |t|< \lambda\\ \frac{t}{1+\eta}, & \mbox{ if }  |t|\geq\lambda. \end{cases} \label{hybridthfunc}
\end{eqnarray}
%\vspace{-.35in}
%
%\noindent
%This offers adaptive selection and shrinkage.
%The two parameters have separate roles: $\lambda$ controls the threshold value while $\eta$ controls the amount of shrinkage.
%This parametrization (rather than $\lambda_1 1_{\theta\neq0} + \lambda_2 \theta^2$) is advantageous for tuning in applications.
Letting $q\equiv 0$, we obtain a penalty function fusing the hard-penalty and the ridge-penalty
%\vspace{-.15in}
\begin{eqnarray}
P_{HR}(\theta;\lambda, \eta)=\begin{cases} -\frac{1}{2} \theta^2 + \lambda |\theta|, &\mbox{ if } |\theta| < \frac{\lambda}{1+\eta}\\ \frac{1}{2} \eta \theta^2 +\frac{1}{2}\frac{\lambda^2}{1+\eta}, &\mbox{ if } |\theta| \geq \frac{\lambda}{1+\eta}. \end{cases} \label{hybridpenfunc}
\end{eqnarray}
%\vspace{-.3in}
%
%\noindent
Moreover,  for $q(\theta;\lambda, \eta)=\frac{1+\eta}{2} (|\theta|-\lambda)^2  1_{0 < |\theta| < \lambda}$, we obtain the $l_0+l_2$ penalty
%\vspace{-.15in}
\begin{eqnarray}
P(\theta)=\frac{1}{2} \eta \theta^2  + \frac{1}{2}\frac{\lambda^2}{1+\eta} 1_{\theta\neq0}. \label{hybridpenfunc2}
\end{eqnarray}
%\vspace{-.45in}
%
%\noindent
%%%See Figure \ref{fighybrid} for an illustration.
%Statistical modeling usually has joint concerns of accuracy and sparsity  interplaying with each other during the fitting.
This hard-ridge penalty offers both selection and shrinkage into regularization, interplaying with each other during the iteration for nonorthogonal designs. In the group situation, the algorithm  aims for a penalty of form $\sum_{k=1}^K \frac{\lambda_k^2}{2(1+\eta_k)} 1_{\|\bsbb_k\| \neq 0}+ \sum_{k=1}^K \frac{\eta_k}{2} \|\bsbb_k\|_2^2$ which is able to deal with collinearity in the design in the pursuit of between-group sparsity.
%, and their iterative interaction results in different sparsity patterns than Hard-TISP. %%temp%% and is able to adapt to different noise levels and sparsities.???
\end{example}
\section{Algorithm Design and Fast Computation}
%\vspace{-.15in}
\subsection{Algorithm design details}
\label{subsec:impl}
%\vspace{-.05in}
Either the global scaling of $\bsbX$ based on ${\mathcal L}_{\Theta}$ or the iteration \eqref{tisp-gen} is simple to implement.  We give more algorithm design details as follows.

First, the  range of the threshold  parameter is finite and can be determined from \eqref{thetaestglm-gen}. Assuming  $\lambda$ is the threshold and $\bsbX$ has been column normalized, we can let $\lambda$ vary over the interval  $0$ to $\|\bsbX^T\bsby-
\bsbX^T\bsbmu(\bsb{0})\|_{\infty}$.

The termination criterion can be  based on $\bsbb^{(j)}$ or $F(\bsbb^{(j)})$.
Extensive simulation studies showed that the approximate solution $\bsbb^{(j)}$ often had good enough performance as $j$ is reasonably large. Each iteration involves only low-cost operations like  matrix-vector multiplications.
Setting a maximum number of iterations can provide a  tradeoff between performance and computational complexity.
Moreover, it can be shown (proof omitted) that for hard-thresholding or hard-ridge thresholding, the limiting $\bsbb^{(\infty)}$ is an ML estimate  or a ridge  estimate, restricted to the selected dimensions. This fact can be used in implementation when the maximum number of iterations allowed  has been reached.

It remains to specify the starting point for any given $\lambda$.
Our theory guarantees local optimality  given any initial point $\bsbb^{(0)}$. One can try multiple random starts in computation, but a nice fact is that  pursuing  the globally optimal solution to \eqref{oriprob-gen} is not at all needed to  achieve significant performance gains over the $l_1$ technique. We have found simply using the zero start, i.e., $\bsbb^{(0)}=\bsb{0}$, makes a   good  choice empirically. It finds a  $\Theta$-estimate close to zero in building a parsimonious model. (Of course, other initializations are available -- see, e.g., \cite{dcprog}.)
%(The excellent experimental results in Section \ref{sec:chtheta} and Section \ref{sec:app} will support such  zero initialization.)
Note that  since  the solution path associated with a nonconvex penalty is generally discontinuous in $\lambda$ for \emph{nonorthogonal} models, even though the penalty and threshold function are differentiable to any order on $(0, +\infty)$, such as the transformed $l_1$~\citep{transfL1},  a pathwise algorithm with warm starts  is easy to trap into poor local optima.
Warm starts for a grid of values of $\lambda$  is not recommended over the zero start unless the problem is convex.

%%\emph{TISP with shift.}
%First, we need to specify how to handle a shift vector $\bsba$ appearing  in the model but not penalized in \eqref{oriprob-gen}, which  may increase the computational cost significantly. %, which  arises in detecting the mean shift outliers.
%This occurs when  an intercept term is included in the penalized GLM. Although for a Gaussian model one can center both $\bsbX$ and $\bsby$ to make the intercept vanish,  in general, centering the response may violate the distribution assumption for nonGaussian GLMs.
%%In many applications, such as  detecting GLM outliers for large-$p$ data,  $\bsba$ is also unknown.
%The alternative optimization can be used: given $\bsbb$, we can run Newton's algorithm to solve for $\bsba$, and given $\bsba$, $\bsbb$ is still updated according to \eqref{tisp-gen} but with the mean vector $\bsbmu(\bsbb, \bsba)=[g^{-1}(\bsbx_i^T\bsbb+\alpha_i)]_{n\times 1}$. Yet the following weighted form of TISP can be used which is often more efficient.
%
%\emph{TISP with weights.}

Finally, the $\lambda_k$ in \eqref{tisp-gen} are not necessarily equal to each other. The regularization vector $\bsb{\lambda}$ can be  component-specific  to offer relative weights in regularizing  the coefficients.
%i.e., $\bsb{\lambda}=[\lambda_1, \cdots, \lambda_p]$, and our conclusions carry over. %If $\bsbX=[\tilde\bsbx_1,\ldots,\tilde\bsbx_p]$ is not column-normalized, we may set $\lambda_i=\lambda \|\tilde\bsbx_i\|_2$.
This weighted form can  handle  GLMs with \emph{dispersion}: $f(y_i;\theta_i, \phi)=\exp[(y_i\theta_i-b(\theta_i))/(A_i\phi) + c(y_i,\phi)]$, where $\phi$ is a dispersion parameter {orthogonal} to  $\theta_i$, and $A_i$ is a known prior weight~\citep{agresti}.
%The dispersion parameter $\phi$ is a nuisance parameter {orthogonal} to  $\theta_i$.
The normal and binomial GLMs are concrete examples. Introducing weights is  also useful when a shift vector $\bsba$ appears  in the model but is {unpenalized}. Two examples are mean-shift outlier detection \citep{SheIPOD} and  the intercept estimation.
 %This occurs when  an intercept term is included in the penalized GLM.
(Note that although one can center both $\bsbX$ and $\bsby$ in  a Gaussian model  to make the intercept vanish,  centering the response may violate the distribution assumption for nonGaussian GLMs.)

\subsection{Fast Computation}
\label{subsec:fastcomp}
%\vspace{-.25in}
%\subsection{Numerical techniques}
%\label{subsec:ext}
%\vspace{-.05in}
The iteration of the proposed algorithm involves no high-complexity operations like matrix inversion. We aim to improve its convergence speed especially for high-dimensional computation.
%We introduce some extensions to the vanilla TISP.

\emph{Numerical techniques.}
Although \eqref{tisp-gen} is a nonlinear process, \textbf{relaxation} and \textbf{asynchronous updating} can be incorporated  to accelerate the convergence. %%to be back%%~\citep{Shethesis}.%%
The asynchronous updating of  \eqref{tisp-gen} leads to  in-place computation of $\bsbb$, and the mean vector $\bsbmu$ is always calculated using the recently updated $\bsbb$. Under the assumptions that $y_i$ are Gaussian and the penalty is convex,  this exactly corresponds to   the \emph{coordinate descent algorithm} in~\citet{FHT}. %As mentioned in Example \ref{ex:bin}, in the GLM setup, ~\citet{fhtglm} but the convergence conditions are not clear.
Yet for nonGaussian GLMs,  experience shows that the original synchronous form seems to be more efficient. %, which might have something to do??? with the fact that the mean vector $\bsbmu$ needs to be re-computed when updating each component of $\bsbb$.
The relaxation of \eqref{tisp-gen} is introduced as
%\vspace{-.15in}
\begin{eqnarray}
\begin{split}
\bsbxi^{(j+1)} &=  (1-\omega)\bsbxi^{(j)} + \omega (\bsbb^{(j)}+\bsbX^T\bsby-\bsbX^T \bsbmu(\bsbb^{(j)})), \\
\bsbb_k^{(j+1)}&= \vec\Theta_k(\bsbxi_k^{(j+1)};{\lambda_k}),  \quad 1\leq k \leq K.
%\vspace{-.2in}
\end{split}
\label{relax}
\end{eqnarray}
%\vspace{-.35in}
%
%\noindent
%%to be back%%which corresponds to Relaxation (I) in~\citet{Shethesis}.
We  used \eqref{relax} with $\omega=2$ in experiments, where the number of iterations can be reduced by about 40\% in comparison to the original form.

\emph{Iterative quantile screening.} To reduce the computational cost even more dramatically in high dimensions without losing much performance,
probabilistic means must be taken into account apart from  the numerical techniques.
A reasonable idea is to screen the predictors (features) preliminarily before running \eqref{tisp-gen}.
%It is  more effective  for fast computation than the previous numerical techniques.
But for  correlated data applications  much more caution is needed to  (a) avoid too greedy preliminary screenings, and  (b) keep the screening principle consistent with the final model fitting criterion. \
We perform iterative feature screening by running group TISP in a \emph{quantile} fashion:
at each iteration step of \eqref{tisp-gen}, we set a threshold value to have exactly $\alpha n$  nonzero components arise in $\bsbb^{(j+1)}$. Similar to Section \ref{sec:algconv}, we can show the  procedure is associated with the \textit{constrained} form of the optimization problem \eqref{oriprob-gen}.  After convergence, $\alpha n$ candidate predictors are picked. As long as $\alpha$ is reasonably large, all relevant predictors can be maintained with high probability. Under  the sparsity assumption, one can set $\alpha<1$; we have found $\alpha=0.8$ to be safe empirically.   Sparsity-pursuing algorithms converge much faster on the   screened (relatively) large-$n$ data.
% Then apply the original TISP to the reduced-dimension data.
If the model is Gaussian, the first step of the iterative quantile screening corresponds to independence screening \citep{fanlv} based on marginal correlation statistics.
%Assuming the model is Gaussian, SIS or FDR corresponds to the first step if $\bsbb^{(0)}=\bsb{0}$. Independence screenings are based on {marginal} statistics, and thus the feature ranking is  meaningful when the features are marginally unrelated.
%The quantile screening is less greedy  and is nonmarginal  seen from its iterative nature. %TISP screening is less greedy and more efficient than  subsampling methods.
%This screening is consistent with the classifier and

%Recently, ultrahigh dimensional learning is proposed where current optimization packages are computationally too expensive if directly applied. % to large-scale  data such as genome-wide association studies with  millions of SNPs.
%More seriously, the huge number of  features will produce spurious correlations and make it difficult  to select the right model or estimate the coefficients stably.  Feature screening thus must come into play beforehand.
%{Independence screenings} SIS and ISIS~\citep{fanlv,fansamwu} are advocated to be
%performed as a fast but crude way to reduce the dimension  to a   moderate size $\alpha n$ (say $\alpha=0.75$); in the second stage a more sophisticated technique such as the $l_1$-penalty or SCAD-penalty   determines the ultimate significant predictors.

%\vspace{-.1in}
\section{Penalty Comparison}
\label{sec:chtheta}
%\vspace{-.1in}
The  design of the penalty $P$ or the threshold function $\Theta$ is an  important topic in applying penalized log-likelihood estimation into real-world problems. We performed systematic simulation studies to compare difference penalty functions in sparse modeling.
%The SCAD seems to be popular in the literature. The firm shrinkage \eqref{firmth} is also used~\citep{mplus}. However, the unbiasedness for large coefficients might not contribute to statistical modeling especially in large-$p$ problems. With %adaptive shrinkage introduced, we could benefit from the bias-variance tradeoff in predictive learning. In fact, if we had learned the truly relevant covariates, imposing a ridge-penalty would often be appropriate. The joint and adaptive consideration of selection and shrinkage is necessary even if we just focus on variable selection, because the regularization parameter is often  tuned according to  the prediction performance.  Hence the hybrid hard-ridge thresholding seems to be an ideal choice.  Its penalty \eqref{hybridpenfunc2} fuses the $l_0$-norm and $l_2$-norm (squared). In addition to the $l_0$-portion which enforces sparsity, the $l_2$-portion handles collinearity and adapts to different noise levels.
%We carried out extensive simulations to compare different penalizations.
Five methods were studied: LASSO (with calibration), one-step SCAD, the nonconvex $l_0$-penalty, SCAD-penalty, and hard-ridge penalty.  The first two are convex but multi-stage. %The plain lasso gave much worse prediction performance and its results were not reported.
Similar to the idea of the \emph{LARS-OLS hybrid} \citep{Efron}, %for each $\lambda$,
we calibrated the LASSO estimate by fitting an unpenalized likelihood model restricted to its selected predictors.   %, but $\lambda$ is chosen based on the `Lasso-MLE' solution path.
One-step SCAD is an example of the one-step LLA~\citep{ZouLi}  which fits a  weighted LASSO with the weights constructed from the ML estimate and the penalty function. We used the previous tuned LASSO-MLE  as the initial estimate in weight construction which behaves  better  than the ML estimate and applies to $p>n$.
The remaining three nonconvex methods  were all be computed by the proposed algorithm. %Hard-TISP corresponds to the $l_0$-penalized log-likelihood model.
%Although SCAD shrinks moderate coefficients,
Neither  SCAD nor $l_0$ introduces estimation bias for large coefficients.
Hard-ridge penalty does simultaneous selection and shrinkage with  a thresholding parameter $\lambda$ and a ridge parameter $\eta$.
For efficiency, we did not run %an empirical search instead of
a full two-dimensional grid search  when looking for the best parameters. Instead, for each $\eta$ in the grid $\{0.5\eta^*,0.05\eta^*,0.005\eta^*\}$ where $\eta^*$ is the optimal ridge parameter, we find $\lambda(\eta)$ to minimize the validation error;  %, the optimal parameter value denoted by $\lambda_o$.
then for $\lambda$ fixed at the optimal value, we find the best $\eta$  to  minimize the validation error.
%The last search may still change the sparsity pattern of the estimate.
%This simple empirical search only compares four solution paths and is more efficient.  %than a full grid search.

\begin{figure}
\centering
\includegraphics[width=1.2\hsize, height=17cm]{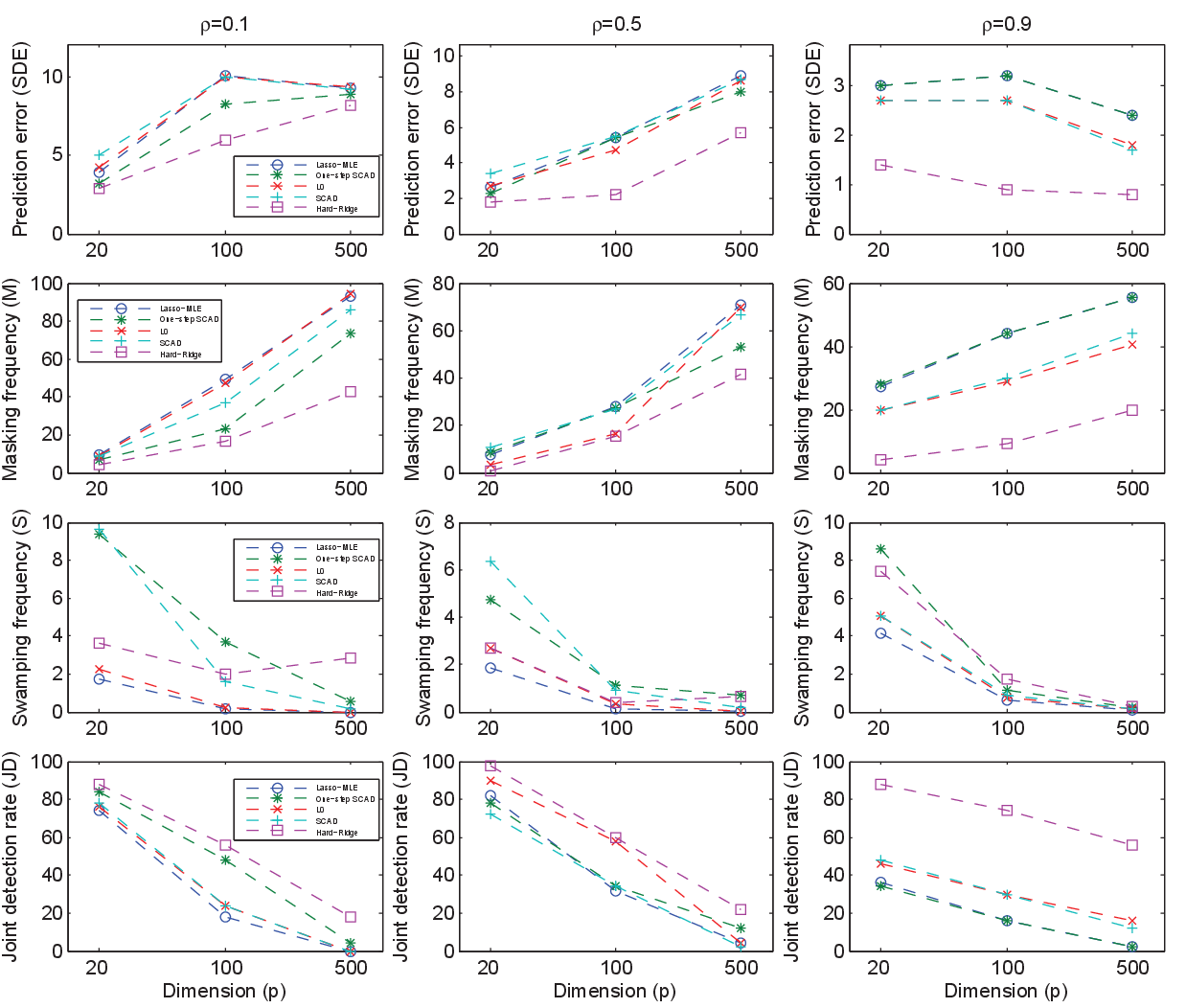}
\caption[]{\small{Performance comparison  of different penalties in terms of
test error, masking/swamping probabilities, and joint identification rate for logistic regression models with $b=0.75$.
%The sample size $n=100$. Note that in variable selection masking is much more serious than swamping, and an ideal method should have $\mbox{SDE} \approx 0$, $\mbox{M}\approx 0\%$, $\mbox{S}\approx 0\%$, $\mbox{ and } \mbox{JD}\approx 100\%$.
}}
\label{figsimu_0.75}
\end{figure}

\begin{figure}
\centering
\includegraphics[width=1.2\hsize, height=17cm]{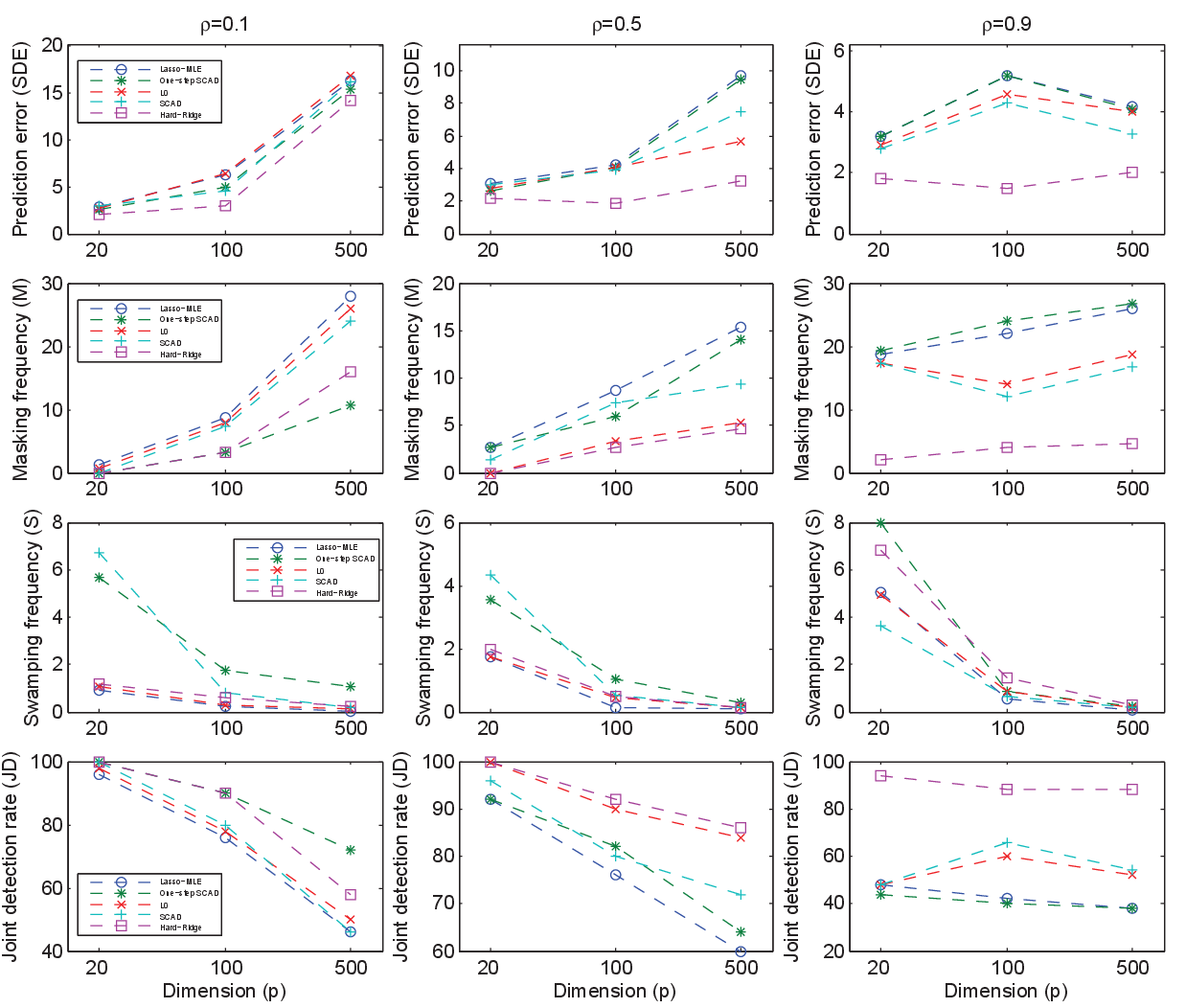}
\caption[]{\small{Performance comparison  of different penalties in terms of
test error, masking/swamping probabilities, and joint identification rate for logistic regression models with $b=1$.
%The sample size $n=100$. Note that in variable selection masking is much more serious than swamping, and an ideal method should have $\mbox{SDE} \approx 0$, $\mbox{M}\approx 0\%$, $\mbox{S}\approx 0\%$, $\mbox{ and } \mbox{JD}\approx 100\%$.
}}
\label{figsimu_1}
\end{figure}

\begin{figure}
\centering
\includegraphics[width=1.2\hsize, height=17cm]{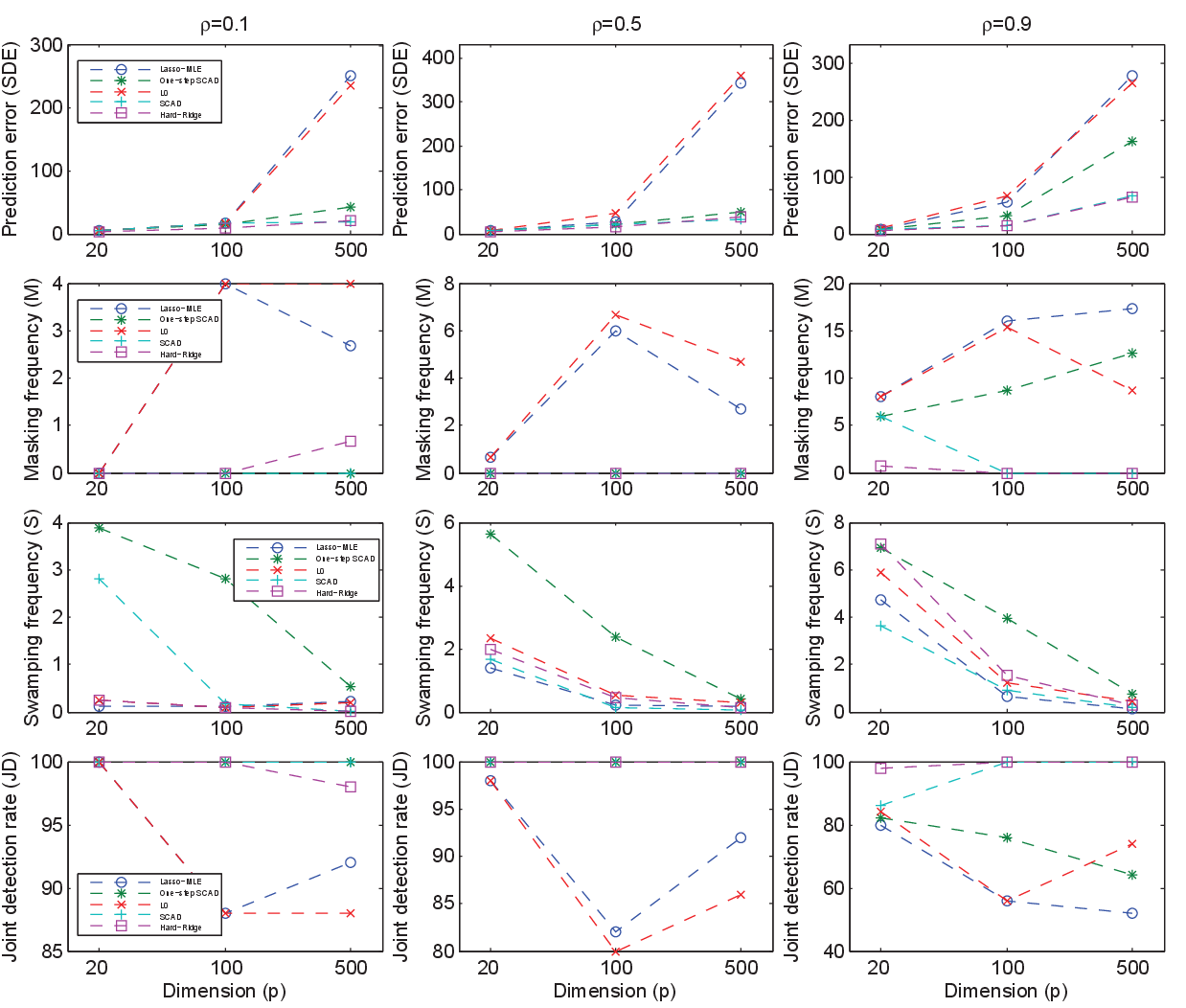}
\caption[]{\small{Performance comparison  of different penalties in terms of
test error, masking/swamping probabilities, and joint identification rate for logistic regression models with $b=2.5$.
%The sample size $n=100$. Note that in variable selection masking is much more serious than swamping, and an ideal method should have $\mbox{SDE} \approx 0$, $\mbox{M}\approx 0\%$, $\mbox{S}\approx 0\%$, $\mbox{ and } \mbox{JD}\approx 100\%$.
}}
\label{figsimu_2.5}
\end{figure}

We seek to  evaluate and compare the performances of different penalties in this section. To understand the true potential of each method in an ideal situation and allow us to draw a stable performance comparison, we tuned all regularization parameters on a very large independent  \textbf{validation} dataset. The simulation setup is as follows.
Let $\bsbb = (b, 0, b, b, 0, \cdots, 0)^T$, $\bsbX=[\bsbx_1, \bsbx_2, \cdots, \bsbx_n]^T$ and $\bsbx_i$ are  i.i.d. $\sim \mbox{MVN}(\bsb{0}, \bsbSig)$ where
$\Sigma_{jk}=\rho^{|j-k|}$, $1\leq j,k\leq p$. %The control parameters are $n$ (sample size), $p$ (dimensionality), $\rho$ (correlation), and $b$ (signal strength).
Note that all group penalties in \eqref{oriprob-gen} use the same $l_2$-norm for within-group penalization. The difference lies in between-group penalties. In the experiment, to make this difference more prominent, we let each  predictor fall  into an individual predictor group.   The control parameters were varied by  $(n,p)=(100,20)$, $(100,100)$, $(100, 500)$, $\rho=0.1,0.5,0.9$, and $b=.75,1,2.5$.
We generated an additional large \textbf{test} dataset with 10,000 observations to evaluate the  performance of any algorithm, as well as an  \textbf{validation} dataset of the same size to tune the regularization parameters.
All $3^3=27$ combinations of the \emph{problem size}, \emph{design correlation}, and \emph{signal strength} were covered in the simulations.
We measured an algorithm's performance  by prediction accuracy and
sparsity recovery, for each model simulated 50 times.
We evaluated the scaled deviance error (SDE) $100 (\sum_{i=1}^N \log f(y_i; \hat\bsbb) /\sum_{i=1}^N \log f(y_i; \bsbb) - 1)$ on the test data. For stability, we reported the $40\%$ trimmed-mean of the SDEs  from the 50 runs.
We also reported  variable selection results  via three benchmark measures:  the mean masking (\textbf{M}) and swamping (\textbf{S}) probabilities, and the rate of successful joint detection (\textbf{JD}). %, and the mean  total misclassification error (MisC).
The masking probability is the fraction of undetected relevant variables (\textit{misses}),
% $100\% \cdot |\{i: \beta_i \neq 0, \hat\beta_i = 0\}|/|\{i: \beta_i \neq 0\}|$,
 the swamping probability is the fraction of spuriously identified variables (\emph{false alarms}),
%$100\% \cdot |\{i: \beta_i = 0, \hat\beta_i\neq 0\}|/|\{i: \beta_i = 0\}|$, which resemble the false nonzero- and zero-identifications  respectively.
and the JD is the fraction of simulations with zero miss. %percentage of the true model nested within the selected.
%defined by 100\% $\times$  the mean  $1_{O\subset \hat O}$ over all simulations, where $O=\{i: \beta_i\neq 0\}$ and $\hat O=\{i: \hat\beta_i\neq 0\}$.
%And the total misclassification error % identification error
%is    $|\{i:  1_{\hat\beta_i}\neq 1_{\beta_i}\}|$. %, which represents the total number of inconsistent zeros and nonzeros for each estimate compared to the true $\bsbg$, and
In variable selection, masking is a much more serious problem than swamping, and an ideal method should have $\mbox{M}\approx 0\%, \mbox{S}\approx 0\%, \mbox{ and } \mbox{JD}\approx 100\%$. %, \mbox{MisC}\approx 0$.
%The simulation results for logistic regression  are summarized in Tables \ref{simutable.75}, \ref{simutable1} and \ref{simutable2.5}. (We also did experiments for Poisson loglinear models which led to similar conclusions. The results are not presented  .)
%Due to  limited space, only the results of the logistic models with $b=0.75$ in Figure \ref{figsimu_0.75}.
The simulation results for logistic regression  are summarized in Figures \ref{figsimu_0.75}, \ref{figsimu_1}, and \ref{figsimu_2.5}. %in Tables \ref{figsimu_0.75}, \ref{figsimu_1} and \ref{figsimu_2.5}.

%The results of the logistic models have been summarized into three figures and presented in the supplemental materials. %in Figures in Figure \ref{figsimu_0.75}, \ref{figsimu_1}, and \ref{figsimu_2.5}. %

We briefly summarize the conclusions as follows. Seen from the results, the Lasso-MLE that chooses $\lambda$ according to the  bias corrected lasso   alleviated the issue that even when the signal-to-noise ratio is pretty high, the lasso overselects~\citep{leng}, but still leaves much room for improvement. The nonconvex $l_0$  yields a restricted ML estimate, too, but is single-stage, and often did better in variable selection. %It also has relatively more power to deal with correlated designs.
%However, shrinking large coefficients is necessary (see, for example, $b=2.5$ and $p=500$). %; the comparisons suggest the benefits of shrinking in prediction as well as in model selection.
The  weighting technique in \textit{one-step} SCAD,  though theoretically effective for $p$ fixed and $n\rightarrow \infty$, requires a careful choice of the initial estimate in finite samples.  %We constructed the SCAD-weights from the tuned `Lasso-MLE', but
The improvement brought by weighting was somewhat limited, especially when some predictors are correlated. % (especially for correlated designs).
Fully solving the nonconvex SCAD problem, though using a na\"{i}ve zero start,  %clearly outperforms the multi-stage methods in this situation and
showed good large-$p$ performance.
%The above methods or penalties are %state-of-the-art
%popular in the literature; overall we can not draw a uniform conclusion of which is the best. On the other hand,
In the 27 experiments, the nonconvex hard-ridge penalization \eqref{hybridpenfunc} (or \eqref{hybridpenfunc2}) had striking advantage in prediction and sparsity recovery simultaneously, in various challenging situations of large $p$, low signal strength, and/or high collinearity. %, and is clearly the estimator of choice.
%Please refer to the supplemental materials for the detailed results and figures.
Its $l_2$-portion dealt with collinearity well and adapted to different noise levels; meanwhile, its $l_0$-portion,  nondifferentiable at zero, enforced higher level of sparsity  than  convex techniques.
%The hybrid hard-ridge thresholding seems to be an ideal choice.
Not surprisingly, in computation, nonconvex penalties  required more computational time than the $l_1$, but the cost is acceptable. For example, in the setup of $(n, p)=(100, 500)$, $\rho=0.5$, $b=0.75$, the total running time (in seconds)  was 322.6, 1184.1, 1910.5, 1553.3 for $l_1$, $l_0$, SCAD, and hard-ridge, respectively. The performance boost, with some  sacrifice in computational time, is affordable.

%*************** soft ***************
%Elapsed time is 322.611042 seconds.
%*************** one step scad ********
%Elapsed time is 2239.718481 seconds.
%*************** hard *****************
%Elapsed time is 1184.083158 seconds.
%**************** scad ****************
%Elapsed time is 1910.487358 seconds.
%***************** hr *****************
%Elapsed time is 1553.266310 seconds.

%\vspace{-.1in}
\section{Choice of the Regularization Parameter}
%\vspace{-.1in}
Parameter tuning plays an important role in penalized log-likelihood estimation.
If we  assume  $\bsbb$ is sparse and  the sample size $n$ is large relative to the true dimensionality (denoted by, say,  $p_{nz}$), then BIC can be used but may still suffer from overselection \citep{chenchen}. % proposed large-$p$  variants of BIC.
Directly cross-validating (CV) the regularization parameter $\lambda$ is also popular in the literature. However, it may not be appropriate for {nonconvex} penalties. (i) The optimal value of $\lambda$ in  the penalized criterion \eqref{oriprob-gen} %, of Lagrangian form)
is a function of the true $\bsbb$ \emph{and} the data $(\bsbX, \bsby)$.  As the training data  change, the optimal value of the penalty parameter may not remain the same. But $K$-fold CV requires $K$ different trainings. (ii)  Even if a nonconvex penalty and its corresponding threshold function are smooth on $(0, \infty)$, the solution path $\hat\bsbb(\lambda)$ is typically discontinuous in $\lambda$ for nonorthogonal designs.  %although $K$-fold CV is widely used to tune the parameters in SCAD and  $l_p$-penalties,
As a consequence, for any given value of $\lambda$, the $K$ fitted  models in CV may not be directly comparable, and thus averaging the CV errors can be unstable and misleading. A crucial question is  how to guarantee  the $K$ trainings (and validations) are associated with  the same model.

To address the issue in sparsity problems,  we propose  $K$-fold  \emph{selective cross-validation} (SCV)  outlined below. Let  $\mathcal{A}$ be a given sparsity algorithm.
\begin{compactenum}
\item   Run $\mathcal A$  on the \emph{whole} dataset for $\lambda$ in a grid of values, getting the solution path $\hat\bsbb_l$, $1\leq l \leq L$. The associated sparsity patterns are denoted by $nz_l=nz(\hat\bsbb_l)$, $1\leq l \leq L$.
\item  For each $l$,  run cross-validation to fit $K$ models with only the predictors picked by $nz_l$. We use \emph{degree-of-freedom (df) matching} to find  proper shrinkage parameter in each training. % the models.
\item Summarize the CV deviance errors and determine the optimal estimate and sparsity pattern in the solution path.
\end{compactenum}
Step 1 determines  candidate sparsity patterns to be used in the training step. Given $k$ ($1\leq k \leq K$), on the data without the $k$th subset, Step 2 fits models \emph{restricted} to the selected dimensions only.
Specifically, if $\hat\bsbb_l$ is from the %$l_1$ or
$l_0$ penalization, the df  is essentially  the number of nonzero components in $\hat\bsbb_l$. % \citep{Zoudof}.
Therefore, in each CV training, we simply fit a model with  MLE,  unpenalized and restricted to $nz_l$.
For the hard-ridge penalty, the contribution of the ridge parameter $\eta$ must be considered, for the df of an  $l_2$-penalized GLM with estimate $\hat\bsbb$ is approximately $Tr\{(\bsbmI(\hat\bsbb) +\eta \bsbI)^{-1} \bsbmI(\hat\bsbb)\}$ \citep{agresti}.
%in matching the $K$ local estimates to the global estimate. Similarly, it is not reasonable to use the same value of $\eta$ in $K$ trainings.
%The complexity of a  $l_2$-penalized GLM estimate $\hat\bsbb$ has degrees of freedom (df) $\mbox{df}(\hat\bsbb, \eta)\approx Tr\{(\bsbmI(\hat\bsbb) +\eta \bsbI)^{-1} \bsbmI(\hat\bsbb)\}$.
%For any $l$,  in the $k$th model fitting without the $k$th subset of data, we choose $\eta$ such that  the restricted penalized estimate has the same df as the $\hat\bsbb_l$. This can be done by bisection search.
To guarantee the $k$th trained ridge model  has the same df as $\hat\bsbb_l$, bisection search can be used to find the appropriate value $\eta_k$.
Finally, in Step 3,   the prediction errors on the left-out piece of data can be summarized by   $-2\sum_{i=1}^n  \log f(y_i; \hat\bsbb_l^{-k(i)})=:\mbox{SCV}(l)$, where $\hat\bsbb_l^{-k(i)}$ denotes the above local  estimate without the $k(i)$th subset and  restricted to the selected dimensions. % and having the same df as $\hat\bsbb_l$.
%$k(i)$ denotes the subset index of observation $i$, and $\hat\bsbb_l^{-k(i)}$ denotes the estimate without the $k(i)$th subset, restricted to the dimensions $nz_l$ and having the same df as $\hat\bsbb_l$.
If the model is very sparse---$p_{nz}\ll n$ and $p_{nz}\ll p$, a BIC correction term can be added:  $\mbox{SCV-BIC}(l)=  \mbox{SCV}(l) + \log n\cdot \mbox{df}(\hat\bsbb_l)$. Empirically, this new criterion can overcome the overselection issue of BIC, through replacing the training error by the SCV error. % (see Section \ref{sec:app}).
%Unlike   Chen \& Chen's EBIC,  SCV-BIC is not affected by appending zero predictor columns to the model matrix.
A similar idea is used in \cite{ccBB}.  

In summary,  SCV runs the  given sparse algorithm only \emph{once} and \emph{globally}, instead of $K$ times locally,  %in the SCV procedure
to determine the common sparsity patterns.  % such that the $K$ trainings refer to the same model.
It can  reduce the computational cost and resolve the model inconsistency issue of the plain CV.  %Implicitly, the $K$ fitted models use different  $\lambda$-values to match the degrees of freedom.

\section{Applications}
%\vspace{-.1in}
\label{sec:app}
We demonstrate the efficacy of  our algorithm for computing  nonconvex penalized models  by  super-resolution spectral analysis in signal processing, and cancer classification and gene selection in microarray data analysis.

\subsection{Super-resolution spectral analysis}
The problem of spectral estimation studies how the signal power is distributed over frequencies, and has rich applications in speech coding and radar sonar signal processing. It becomes very challenging when the required frequency resolution is high, because the number of the frequency levels at a desired resolution can be (much) greater than the sample size, referred to as \emph{super-resolution} spectral estimation.
Super-resolution spectral analysis goes beyond the traditional Fourier analysis and is one of the first areas where the $l_1$-relaxation technique, i.e., the \emph{Basis Pursuit} by \citet{basispursuit}, was proposed. Here we revisit the problem and demonstrate the advantage brought by group nonconvex  penalized likelihood estimation. We focus on the classical \textbf{TwinSine} signal arising from target detection:
$$
y(t) = a_1\cos(2\pi f_1t+\phi_1)+a_2\cos(2\pi f_2t+\phi_2) + n(t)
$$
where $a_1=2$, $a_2=3$, $\phi_1=\pi/3$, $\phi_2=\pi/5$, $f_1=0.25$Hz, $f_2=0.252$Hz and $n(t)$ is white Gaussian noise with
variance $\sigma^2$. Obviously, the frequency resolution needs to be as fine as $0.002$ Hz to perceive and distinguish the two sinusoidal components.
For convenience, assume the data sequence is evenly sampled at  $n=100$ time points $t_i=i$, $1\leq i \leq n$. (Our approach  does {not} require uniform sampling.)
An  \emph{overcomplete} dictionary  to attain the desired frequency resolution can be constructed by setting the maximum frequency $f_{\max}=1/2=0.5$ Hz, and the number of frequency bins $D=250$. Concretely, let $f_k = f_{\max} \cdot k/D$ for $k=0, 1, \cdots, D$ and define the frequency atoms $\bsbX_{\cos} =[\cos(2\pi t_i  f_k)]_{1\leq i \leq n, 1\leq k \leq D}$ and $\bsbX_{\sin} =[\sin(2\pi t_i  f_k)]_{1\leq i \leq n, 1\leq k \leq D-1}$, where the last sine atom vanishes because $\sin(2\pi t_i  f_D)=0$ for integer-valued $t_i$.
Then $\bsbX = [\bsbX_{\cos}\  \bsbX_{\sin}]$ is of dimension 100-by-499 without the intercept, resulting in a challenging high-dimensional learning problem.
In this situation, the classical Fourier transform based periodogram or least-squares periodogram (LSP) suffers from severe power leakage, while the basis pursuit (BP) is able to   super-resolve under the spectral  sparsity assumption.
On the other hand,
the \emph{pairing} structure of cosine and sine atoms is often ignored in spectrum recovery.
More seriously, when the desired frequency resolution is sufficiently  high,
the dictionary contains  many similar sinusoidal components and the high pairwise correlations may make  the $l_1$ relaxation of the $l_0$-norm corrupted  in selecting all frequencies consistently.

We simulated the signal model at  given noise levels $\sigma^2=8,1,0.1$, each with 20 times
to evaluate the performance of an algorithm.
At each run, we generated  additional test data at $N=2000$ time points different than those of the training data to calculate the effective prediction error   $\mbox{MSE}^*=\sum_{i=1}^{N} (y_i - \bsbx_i^T \hat \bsbb-\hat\alpha)^2/N-\sigma^2$. The median of $\mbox{MSE}^*$  was reported, denoted by \textbf{Err}, as the goodness of fit of the obtained model.
The frequency detection is measured by joint detection rates -- \textbf{JD}, misses -- \textbf{M}, and false alarms -- \textbf{S} defined in Section \ref{sec:chtheta}.
Table~\ref{tab:spec} compares the performance of BP, grouped lasso, hard-ridge and grouped hard-ridge penalized regressions on the TwinSine signal, all of which were computed via the proposed algorithm. %\eqref{tisp-gen}.

\begin{table}[ht]
\centering
\setlength{\tabcolsep}{1mm}
\caption{\small{Performance comparison of  basis pursuit, grouped lasso (G-Lasso), hard-ridge and grouped hard-ridge (G-Hard-Ridge) penalized regressions for spectral estimation.} }\label{tab:spec}

\small{
\begin{tabular}{l c c rrrr rrrr rrrr}
\hline\hline
& & &
\multicolumn{4}{c}{$\sigma^2 = 8$\ \  }&
\multicolumn{4}{c}{$\sigma^2 = 1$\ \  }&
\multicolumn{4}{c}{$\sigma^2 = 0.1$}\\
 & Tuning & &
\multicolumn{4}{c}{$\mbox{SNR}=18.13$\ \  }&
\multicolumn{4}{c}{$\mbox{SNR}=8.13$\ \  }&
\multicolumn{4}{c}{$\mbox{SNR}=-0.90$}\\
%\hline
\cline{4-15}
& & &
{Err }&{JD}&{M}&{ S\ \  } &
{Err }&{JD}&{M}&{ S\ \  } &
{Err }&{JD}&{M}&{ S}\\
\hline
{Basis pursuit} & {Large-Val} & &
 4.15 & 0 & 55 & 0.5  \  \  & 3.00 & 0.0  & 50 & 0.3 \ \   &  2.87 & 0.0  & 50 & 0.3  \\
Hard-Ridge & {Large-Val} & &
 1.70 & 45 & 16.3 & 0.4 \ \   & 0.36  & 80 & 5 & 0.2 \ \   & 0.03 & 100 & 0 & 0.1  \\
%\hline
%\hdashline
\hdashline[1pt/2pt]
G-LASSO & {Large-Val} & &
 1.58 & 90 & 5.0 & 1.8  \ \   & 0.25  & 100 & 0 & 1.6 \ \   &  0.04 & 100 & 0 & 2.8 \\
\textbf{G-Hard-Ridge} & {Large-Val} & &
 0.66 & 95 & 2.5 & 0.1  \ \   &0.16  & 100 & 0 & 0.0 \ \   &  0.02 & 100 & 0 & 0.0 \\
\hline
\textbf{G-Hard-Ridge} & \emph{\textbf{SCV-BIC}} & &
 1.11  & 85 & 7.5 & 0.0  \ \   &0.27  & 100 & 0 & 0.0 \ \   &  0.12 & 100 & 0 & 0 \\
\hline\hline
\end{tabular}
}
\end{table}						

To see the true potential of each penalty in an ideal situation, in the first 4 experiments we used independent  large validation data  (of 2000 observations) to tune the parameters. The penalty comparison showed the improvement of the nonconvex  hard-ridge penalty % over the popular $l_1$-penalty
in both time-domain prediction  and frequency-domain spectrum reconstruction.
In the last experiment,  the hard-ridge was run with no additional validation data.  %only the training data (of 100 observations) used for parameter tuning.
We used SCV with BIC correction on the 100 training observations.

Again, our results  showed that pursuing the global minimum of the nonconvex criterion \eqref{oriprob-gen} is not necessary; the zero start in \eqref{tisp-gen}  offered good accuracy and regularization. %As expected, the computational complexity is higher than the convex $l_1$ but is acceptable.
In the experiments we predefined a maximum iteration number  $M_{\max}=5000$ (see Section \ref{subsec:impl}). To solve the $l_0+l_2$ type problems, our algorithm required 4 to 6 times as much time as the $l_1$ in computing one solution path.
The higher computational complexity   is expected but is an acceptable tradeoff between performance and computational complexity in super-resolution spectral analysis.

%The tuned hard-ridge (of group form) achieved excellent  performance in super-resolution spectral estimation.

\subsection{Classification and gene selection}
We then illustrate our algorithms with an example of cancer classification with joint gene selection. We analyzed  real acute lymphoblastic leukemia (ALL) data conducted with HG-U95Av2 Affymetrix arrays~\citep{chia}.
Following~\citet{Schol}, we focus on the B-cell samples and would like to contrast the patients with the BCR/ABL
fusion gene resulting from a translocation of the chromosomes 9 and 22, with those who are cytogenetically normal (NEG).
The preprocessed data can be loaded from the Bioconductor data package {\tt ALL}.  %the code for probeset filtering is available in~\citet{Schol}.
This leads to 2,391 probe sets and 79 samples, 42 labeled with ``NEG" and 37 labeled with ``BCR/ABL".

We first ran iterative quantile screening introduced in Section \ref{subsec:fastcomp} for dimension reduction. %to remove some irrelevant genes.
%We used the hard-ridge $\Theta$.
Specifically, we ran quantile TISP
with the hard-ridge thresholding function for $\eta$ in a small grid
of values, and then  chose the optimal one by 5-fold SCV.
We used $\alpha=0.8$. % and obtained $\alpha n$ candidate genes. %, making the problem simpler.
Then  we  ran the original form of the algorithm \eqref{tisp-gen} to solve hard-ridge penalized logistic regression. The parameters were  tuned by 5-fold SCV with no/AIC/BIC correction. %  on the reduced-dimension microarray data.
For comparison, we tested another two up-to-date classifiers with \emph{joint} gene selection: the nearest shrunken centroids~\citep{nsc} (denoted by NSC) and the Ebay algorithm~\citep{ebay}. (The results of the $l_1$ penalized logistic regression are not reported, because in comparison, it gave similar error rates but selected too many ($>30$) genes.) For an implementation of NSC,  refer to the package
{\tt pamr} in R. The R-code for Ebay is also available online~\citep{ebay}.  Their regularization parameters were tuned by cross-validation.
To prevent from getting over-optimistic error rate estimates, we used a \emph{hierarchical} cross-validation procedure where an outer 10-fold CV was used for performance evaluation while the inner CVs were used for parameter tuning.
Table \ref{tableALL} summarizes the  prediction and selection performances of the three classifiers. % based on 10-fold (outer) CV.
The proposed algorithms had excellent performance. Hard-ridge-penalty with  SCV-BIC tuning behaved the best for the given data: it gave the smallest error rate and produced the most parsimonious model with only about 8 genes involved.

%\vspace{-.15in}

\begin{table}[htp]
\caption{\small{Prediction error and the number of selected genes.}}

\vspace{-.2in}
\begin{center}
\small{
\begin{tabular}{l c c } %{l l c c c c c c c c c}
\hline
%\multicolumn{2}{c}{$L$} & \multicolumn{4}{c}{15}& & \multicolumn{4}{c}{20}\\
%\cline{3-6}  \cline{8-11}

%\multicolumn{2}{c}{$L=-1$, $O$=} &  200 & 100 & 50 & 20 & 10 \\% & & 200 & 100 & 50 & 10 \\
  & { Misclassification error rate }  &  {\# of selected genes  } \\
& (mean, median) & (mean, median) \\
\hline
%\multirow{2}{*}{ }
NSC    & 16.4\%, 12.5\%  & 19.3, 14\\
\hline
Ebay & 12.7\%, 12.5\% & 16.2, 16 \\

\hline
Hard-Ridge with SCV & 11.3\%, 12.5\% & \ 21.4, 22.5 \\
Hard-Ridge with SCV-AIC & 10.2\%, 6.3\% & 11.2, 9.5 \\
Hard-Ridge with SCV-BIC & \ 8.9\%,  6.3\% & 8.1, 8 \\

\hline

\hline
\end{tabular}
}
\end{center}
\label{tableALL}
\end{table}
%}

%\vspace{-.2in}

\begin{figure}[h]
\begin{center}
\includegraphics[width=4.0in, height=2.2in]{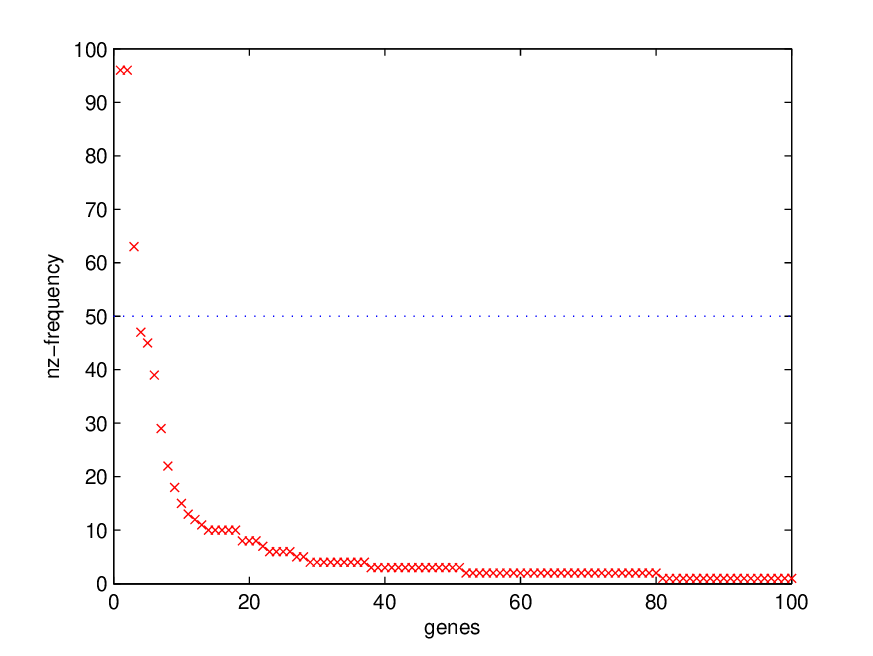}
\hspace{-.37in}
\includegraphics[width=4.0in, height=2.2in]{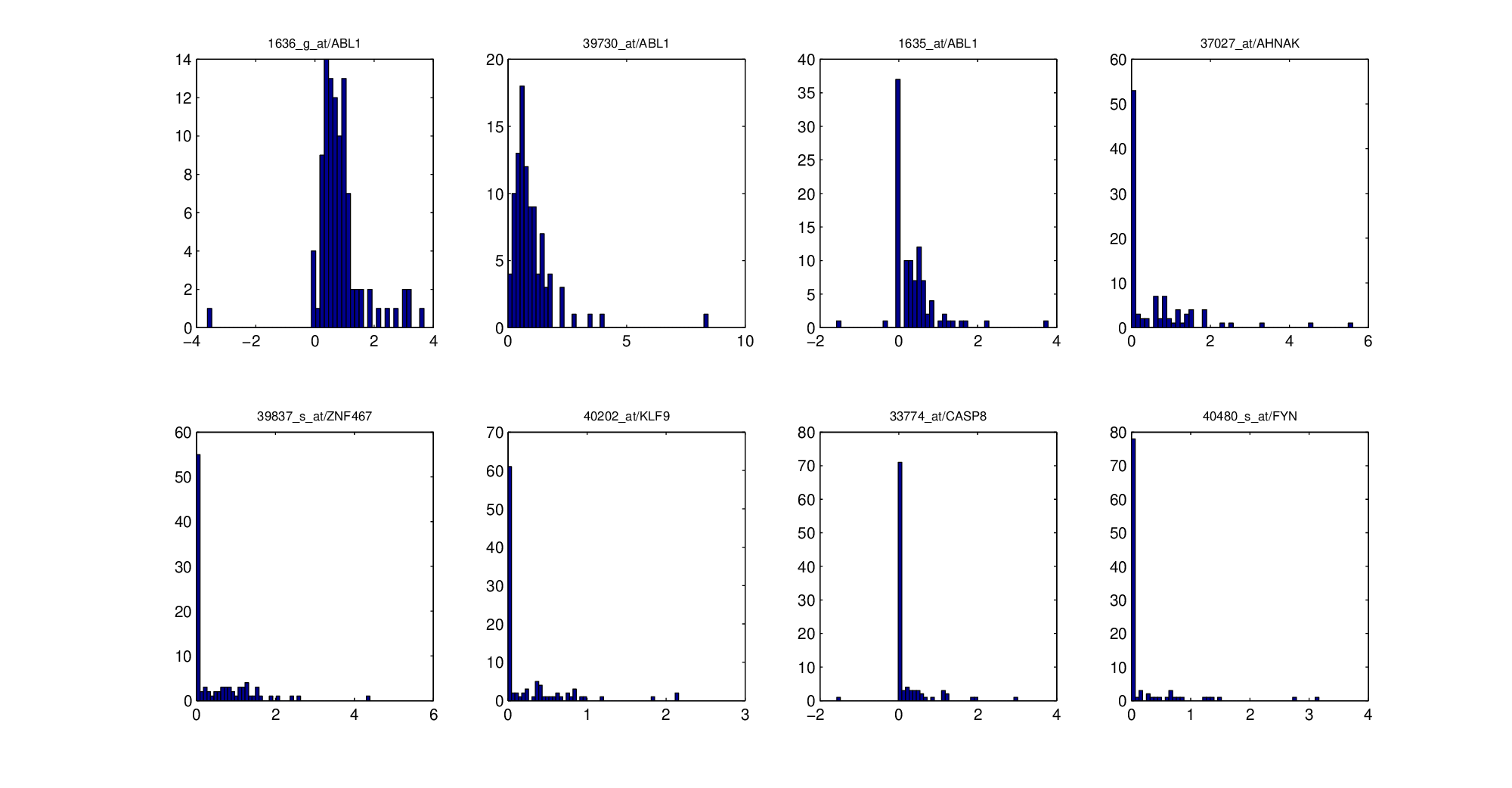}
\end{center}
\vspace{-.3in}
\caption[bootstrap coefficients]{\small{Upper panel: Proportions of the coefficient estimates being nonzero over the $100$ bootstrap replications (only the top 100 genes are plotted). Lower panel: Histograms of the bootstrap coefficient estimates of the top 8 genes.
}}
\label{figprops}
\end{figure}

Next we identify the relevant genes. We bootstrapped the data  $100$ times.
For each bootstrap dataset, after standardizing the predictors, we fit a hard-ridge penalized logistic regression  with the parameters tuned by 5-fold SCV-BIC.
Figure \ref{figprops} plots the frequencies of the coefficient estimates being nonzero  and the estimate histograms over the $100$ replications. % of the top 8 genes. are plotted in Figure \ref{fighists}.
The bootstrap results give us a  confidence measure  of selecting each gene.
The top three probesets had nonzero coefficients more frequently ($>50\%$ of the time) and they jointly appeared  $63$ times in the selected models, the most frequently visited triple in bootstrapping.
Annotation shows that all three probe sets -- 1636\_g\_at, 39730\_at, and 1635\_at -- are associated with the same gene -- ABL1.
%\section{Real Data Example}

%\begin{figure}[h]
%\begin{center}
%%$\begin{array}{c}
%%\epsfxsize=8cm \epsffile{./figs/cooling-double.eps}
%\includegraphics[width=4.5in, height=3.5in]{./figs/nzfreqs-100.eps}
%%\includegraphics[width=8cm, height=6cm]{./figs/cooling-conv.png}
%%\end{array}$
%\end{center}
%\caption[Non-zero Hybrid-TISP coefficient proportions]{\small{Proportions of the Hybrid-TISP coefficients being nonzero over the $100$ bootstrap replications. Only the top 100 genes are plotted.
%}}
%\label{figprops}
%\end{figure}
%
%\begin{figure}[h]
%\begin{center}
%%$\begin{array}{c}
%%\epsfxsize=8cm \epsffile{./figs/cooling-double.eps}
%\includegraphics[width=5in]{./figs/hists.eps}
%%\includegraphics[width=8cm, height=6cm]{./figs/cooling-conv.png}
%%\end{array}$
%\end{center}
%\caption[TISP coefficient histograms]{\small{Histograms of the bootstrap coefficient estimates for the top 8 genes.}}
%\label{fighists}
%\end{figure}

%\vspace{-.1in}
\section{Conclusion}
%\vspace{-.1in}
The paper proposed a simple-to-implement algorithm   for solving penalized log-likelihoods. The predictors can be arbitrarily grouped to pursue the between-group sparsity and %In contrast to~\citet{Yuan},
we do not require the within-group predictors to be orthogonal. Our treatment is rigorous and applies to any GLM.
We proved a convergence condition %which are  stronger than the available results
%~\citet{FHT},~\citet{fhtglm},~\citet{SheTISP}, and~\citet{fhtnote}
in theory and  it leads to a tight preliminary scaling which helps reduce the  number of iterations in implementation. Our algorithm and theoretical analysis  allow  for essentially any nonconvex penalty, and a  $q$-function trick was used to attain the exact discrete $l_0$ and $l_0+l_2$ penalties. % including the (group) $l_0$-penalty. %(See Section \ref{subsec:th} for a comparison of our achievements with some relevant works.)
\appendix
\section{Proof of Theorem \ref{conv-gen}}
\label{app}
%We prove  in the appendix.
%\vspace{-.05in}
\begin{lemma}
\label{uniqsol-gen-grp}
Given an arbitrary thresholding rule $\Theta$,
let $P$ be any function satisfying $P(\theta;\lambda)-P(0;\lambda)=P_{\Theta}(\theta; \lambda) + q(\theta; \lambda)$ where $P_{\Theta}(\theta; \lambda)\triangleq \int_0^{|\theta|} (\sup\{s:\Theta(s;\lambda)\leq u\} - u) \rd u$, $q(\theta; \lambda)$ is nonnegative and $q(\Theta(t;\lambda))=0$ for all $t$.
%let $P$ be any function satisfying $P(\|\bsbb\|_2;\lambda)=P(0;\lambda)+P_{\Theta}(\|\bsbb\|_2; \lambda) + q(\|\bsbb\|_2)$ where $q(\cdot)$ is nonnegative and $q(\|\Theta(\bsbb;\lambda)\|_2)=0$ for all $\bsbb$.
Then, the minimization problem
%\vspace{-.2in}
\begin{align*}
\min_{\bsbb \in {\mathbb R}^n}  \frac{1}{2}\|\bsby-\bsbb\|_2^2 + P(\|\bsbb\|_2;\lambda)\triangleq Q(\bsbb;\lambda)
\end{align*}
%\vspace{-.5in}
%
%\noindent
has a unique optimal solution given by $\hat\bsbb=\vec\Theta(\bsby;\lambda)$ for every $\bsby$ provided that $\Theta(\cdot;\lambda)$ is continuous at $\|\bsby\|_2$.
\end{lemma}
%\vspace{-.1in}
%This is a generalization of Proposition 3.2 in~\citet{antrev}.
\noindent Note that $P$ (and $P_{\Theta}$) may not be differentiable at 0 and may be nonconvex.
For notational simplicity, we simply write $Q(\bsbb)$ for $Q(\bsbb; \lambda)$ when there is no ambiguity. This lemma may be considered as a generalization of Proposition 3.2 in \cite{antrev}.

\noindent \textit{Proof of Lemma \ref{uniqsol-gen-grp}}.
First, it suffices to consider $\bsbb$ satisfying $y_i \beta_i \geq 0$ because for any $\bsbb$, $Q(\bsbb)\geq Q(\bsbb')$ with $\beta_i'= \mbox{sgn}(y_i) |\beta_i|$.
By definition, we have
%\vspace{-.02in}
\begin{eqnarray*}
Q(\bsbb)-Q(\hat\bsbb)&=& -\bsby^T(\bsbb-\hat\bsbb) + \frac{1}{2} (\|\bsbb\|_2^2 - \|\hat\bsbb\|_2^2) +P_{\Theta}(\|\bsbb\|_2; \lambda) - P_{\Theta}(\|\hat\bsbb\|_2;\lambda) \\
&&\qquad + q(\|\bsbb\|_2; \lambda) - q(\Theta(\|\bsby\|_2;\lambda); \lambda)\\
&=& -\bsby^T(\bsbb-\hat\bsbb) + \int_{\Theta(\|\bsby\|_2;\lambda)}^{\|\bsbb\|_2} (u+\Theta^{-1}(u;\lambda)-u) \rd u + q(\|\bsbb\|_2; \lambda).
\end{eqnarray*}
%\vspace{-.5in}
%
%\noindent
On the other hand,
%\vspace{-.05in}
\begin{eqnarray*}
-\bsby^T(\bsbb-\hat\bsbb)&=&-\bsby^T \bsbb+ \|\bsby\|_2 \Theta(\|\bsby\|_2; \lambda)\\
&\geq& -\|\bsby\|_2 \|\bsbb\|_2+ \|\bsby\|_2 \Theta(\|\bsby\|_2; \lambda)\\
&=&  -\|\bsby\|_2(\|\bsbb\|_2-\Theta(\|\bsby\|_2; \lambda)\\
&=& -\|\bsby\|_2(\|\bsbb\|_2- \|\hat\bsbb\|_2).
\end{eqnarray*}
%\vspace{-.55in}
%
%\noindent
Hence
$
Q(\bsbb)-Q(\hat\bsbb) \geq  \int_{\Theta(\|\bsby\|_2;\lambda)}^{\|\bsbb\|_2} (\Theta^{-1}(u;\lambda)-\|\bsby\|_2) \rd u + q(\|\bsbb\|_2; \lambda).
$

Suppose $\|\bsbb\|_2>\Theta(\|\bsby\|_2;\lambda)$. By definition $\Theta^{-1}(\|\bsbb\|_2;\lambda)\geq \|\bsby\|_2$, and thus $Q(\bsbb)\geq Q(\hat\bsbb)$. Furthermore, there must exist some $u\in[\Theta(\|\bsby\|_2;\lambda), \|\bsbb\|_2)$ s.t. $\Theta^{-1}(u;\lambda)>\|\bsby\|_2$, and hence $Q(\bsbb)>Q(\hat\bsbb)$ due  to the monotonicity of $\Theta^{-1}$. In fact, if this were not true,  we would have $\Theta(t;\lambda)>\|\bsbb\|_2\geq\Theta(\|\bsby\|_2;\lambda)$ for any $t>\|\bsby\|_2$, and $\Theta(\cdot; \lambda)$ would be discontinuous at $t$.
A similar reasoning applies to the case when $\|\bsbb\|_2<\Theta(\|\bsby\|_2;\lambda)$. The proof is now complete. \qed

Hereinafter, we always assume $\Theta(t;\lambda)$ is continuous at any $t$ to be thresholded, since a practical thresholding rule usually has at most finitely many discontinuity points and such discontinuities rarely occur in any real application.

%
%\begin{lemma}
%\label{ineq-grp}
%For any $\bsb{a}\neq \bsb{0}$,
%$
%\|\bsb{a} + \bsb{b}\|_2 - \|\bsb{a}\|_2\geq \bsb{b}^T \bsb{a}^{\circ}.
%$
%\end{lemma}
%\noindent \textit{Proof of Lemma \ref{ineq-grp}}.
%By definition, $\bsb{a}^{\circ}=\bsb{a}/\|\bsb{a}\|_2$ and thus $$\bsb{b}^T \bsb{a}^{\circ} + \|\bsb{a}\|_2=\bsb{a}^T(\bsb{a}+\bsb{b})/\|\bsb{a}\|_2.$$ The lemma follows from Cauchy's inequality.
%\qed

\begin{lemma}
\label{unifuncopt-grp}
Let $Q_0(\bsbb)= \|\bsby-\bsbb\|_2^2/2 + P_{\Theta}(\|\bsbb\|_2;\lambda)$. Denote by $\hat\bsbb$  the unique minimizer of $Q_0(\bsbb)$. Then for any $\bsbdelta$,
$
Q_0(\hat\bsbb+\bsbdelta)-Q_0(\hat\bsbb) \geq {C_1} \|\bsbdelta\|_2^2/2,
$
where  $C_1 = \max(0, 1-{\mathcal L}_{\Theta})$.
\end{lemma}
\vspace{-.05in}

\noindent \textit{Proof of Lemma \ref{unifuncopt-grp}}.
Let $s(u;\lambda) =  \Theta^{-1}(u;\lambda)-u=\sup\{t:\Theta(t;\lambda)\leq u\}-u$.
We have
\vspace{-.05in}
\begin{align}
Q_0(\hat\bsbb+\bsbdelta)-Q_0(\hat\bsbb) &= \frac{1}{2}\|\hat\bsbb+\bsbdelta-\bsby\|_2^2 - \frac{1}{2}\|\hat\bsbb-\bsby\|_2^2  + P_{\Theta}(\|\hat\bsbb+\bsbdelta\|_2)-P_{\Theta}(\|\hat\bsbb\|_2)\notag\\
&=\frac{1}{2}\|\bsbdelta\|_2^2  + (\hat\bsbb-\bsby)^T\bsbdelta + \int_{\|\hat\bsbb\|_2}^{\|\hat\bsbb+\bsbdelta\|_2} s(u;\lambda)\rd u \label{qdiff}
\vspace{-.05in}
\end{align}
%\vspace{-.05in}
%
%\noindent
(i)
If $\hat\bsbb=\bsb{0}$,  $\vec\Theta(\bsby;\lambda)=\bsb{0}$ and so $\Theta(\|\bsby\|_2;\lambda)=0$, from which it follows that $\|\bsby\|_2\leq \Theta^{-1}(0;\lambda)$. Therefore,
\vspace{-.05in}
\begin{align*}
(\hat\bsbb-\bsby)^T\bsbdelta \geq -\|\bsby\|_2 \cdot \|\bsbdelta\|_2 \geq -\Theta^{-1}(0;\lambda) \|\bsbdelta\|_2=-\int_{\|\hat\bsbb\|_2}^{\|\hat\bsbb+\bsbdelta\|_2} s(\|\hat\bsbb\|_2;\lambda)\rd u.
\vspace{-.05in}
\end{align*}
%\vspace{-.02in}
(ii)
If $\hat\bsbb\neq \bsb{0}$, it is easy to verify by Lemma \ref{uniqsol-gen-grp} that $\hat\bsbb$ satisfies $\bsby -\hat\bsbb=s(\|\hat\bsbb\|_2;\lambda) \bsby^{\circ}=s(\|\hat\bsbb\|_2;\lambda) \hat\bsbb^{\circ}$, and thus $$(\hat\bsbb-\bsby)^T\bsbdelta=-s(\|\hat\bsbb\|_2;\lambda) \bsbdelta^T \hat\bsbb^{\circ}.$$
For any $\bsb{a}\neq \bsb{0}$, it follows from Cauchy's inequality that $$\bsb{b}^T \bsb{a}^{\circ} + \|\bsb{a}\|_2=\bsb{a}^T(\bsb{a}+\bsb{b})/\|\bsb{a}\|_2 \leq \|\bsb{a} + \bsb{b}\|_2,$$ or
$
\|\bsb{a} + \bsb{b}\|_2 - \|\bsb{a}\|_2\geq \bsb{b}^T \bsb{a}^{\circ}.
$
Making use of this fact,  we obtain
%\vspace{-.15in}
$$
(\hat\bsbb-\bsby)^T\bsbdelta \geq -(\|\bsbdelta +\hat\bsbb\|_2-\|\hat\bsbb\|_2)s(\|\hat\bsbb\|_2;\lambda) = -\int_{\|\hat\bsbb\|_2}^{\|\hat\bsbb+\bsbdelta\|_2} s(\|\hat\bsbb\|_2;\lambda)\rd u.
$$
%\vspace{-.45in}
%
%\noindent
In either case, \eqref{qdiff} can be bounded in the following way:
%\vspace{-.15in}
\begin{align*}
& Q_0(\hat\bsbb+\bsbdelta)-Q_0(\hat\bsbb) \geq \frac{1}{2}\|\bsbdelta\|_2^2 + \int_{\|\hat\bsbb\|_2}^{\|\hat\bsbb+\bsbdelta\|_2} (s(u;\lambda) - s(\|\hat\bsbb\|_2;\lambda))\rd u\\
=& \frac{1}{2}\|\bsbdelta\|_2^2 + \int_{\|\hat\bsbb\|_2}^{\|\hat\bsbb+\bsbdelta\|_2} \left((\Theta^{-1}(u;\lambda) - \Theta^{-1}(\|\hat\bsbb\|_2;\lambda)) - (u-\|\hat\bsbb\|_2)\right)\rd u.
\end{align*}
%\vspace{-.45in}
%
%\noindent
By  the {Lebesgue Differentiation Theorem}, $(\Theta^{-1})'$ exists almost everywhere and
%\vspace{-.05in}
$$\int_{\|\hat\bsbb\|_2}^{\|\hat\bsbb+\bsbdelta\|_2} (\Theta^{-1}(u;\lambda) - \Theta^{-1}(\|\hat\bsbb\|_2;\lambda))\rd u \geq \int_{\|\hat\bsbb\|_2}^{\|\hat\bsbb+\bsbdelta\|_2} \int_{\|\hat\bsbb\|_2}^{u}(\Theta^{-1})'(v;\lambda)\rd v \rd u.$$
%\vspace{-.4in}
%
%\noindent
By the definition of ${\mathcal L}_{\Theta}$,  $ Q_0(\hat\bsbb+\bsbdelta)-Q_0(\hat\bsbb) \geq \frac{1}{2}\|\bsbdelta\|_2^2 -\frac{{\mathcal L}_{\Theta}}{2}(\|\hat\bsbb+ \bsbdelta\|_2-\|\hat\bsbb\|_2)^2 $. Lemma \ref{unifuncopt-grp} is now proved. \qed

Now we prove the theorem. Recall that the model matrix is $\bsbX=\left[\bsbx_1, \bsbx_2, \cdots, \bsbx_n\right]^T=
[\bsbX_1, \cdots, \bsbX_K]\in {\mathbb R}^{n\times p}$.
%As an alternative to $F(\bsbb)$ given by \eqref{oriprob-gen}, define
Define
%\vspace{-.15in}
\begin{eqnarray}
G(\bsbb,\bsbg)&=& -\sum_{i=1}^n L_i(\bsbg) +\sum_{k=1}^K P_{k}(\|\bsbg_k\|_2; {\lambda}_{k})+ \frac{1}{2} \| \bsbg-\bsbb\|_2^2 \notag \\
&&- \sum_{i=1}^n (b(\bsbx_i^T \bsbg) - b(\bsbx_i^T\bsbb)) + \sum_{i=1}^n \mu_i(\bsbb) (\bsbx_i^T\bsbg-\bsbx_i^T\bsbb).
\label{gdef}
\end{eqnarray}
%\vspace{-.2in}
%
%\noindent
Given $\bsbb$,  algebraic manipulations (details omitted) show that minimizing $G$ over $\bsbg$ is equivalent to
%\vspace{-.15in}
\begin{eqnarray}
\min_\bsbg \frac{1}{2} \left\|\bsbg - \left [\bsbb+\bsbX^T\bsby-\bsbX^T\bsbmu(\bsbb)\right ]\right\|_2^2 +\sum_{k=1}^K P_{k}(\|\bsbg_k\|_2; {\lambda}_{k}). \label{optovergamma}
\end{eqnarray}
%\vspace{-.1in}
%
%\noindent
By Lemma \ref{uniqsol-gen-grp}, the unique optimal solution can be obtained through multivariate thresholding
%\vspace{-.05in}
$$
\bsbg_k=\vec\Theta_k(\bsbb_k+\bsbX_k^T\bsby-\bsbX_k^T\bsbmu(\bsbb); \lambda_k), \quad 1\leq k \leq K
$$
%\vspace{-.35in}
%
%\noindent
even though $P_k$ may be nonconvex.
This indicates the  iterates defined by \eqref{tisp-gen} can be characterized by $\bsbb^{(j+1)} = \arg \min_\bsbg G(\bsbb^{(j)}, \bsbg)$.
Furthermore, for any $\bsbdelta \in {\mathbb R}^p$ we obtain
%\vspace{-.15in}
\begin{align}
G(\bsbb^{(j)}, \bsbb^{(j+1)}+\bsbdelta) - G(\bsbb^{(j)}, \bsbb^{(j+1)})\geq \frac{C_1'}{2} \|\bsbdelta\|_2^2 +\sum_k q_k(\|\bsbb_k^{(j+1)}+\bsbdelta_k\|_2; \lambda_k),  \label{gbound}
\end{align}
%\vspace{-.45in}
%
%\noindent
where $C_1'=\max(0, 1-\max_k {\mathcal L}_{\Theta_k})$, %\max(0, 1-L_{\Theta_1, \cdots, \Theta_K})$,
by applying  Lemma \ref{unifuncopt-grp}, and noting that $q_k (\|\bsbb_k^{(j+1)}\|_2; \lambda_k )=0$ by definition.
Taylor series expansion gives
%\vspace{-.15in}
\begin{eqnarray*}
\sum_{i=1}^n (b(\bsbx_i^T \bsbb^{(j+1)}) - b(\bsbx_i^T\bsbb^{(j)})) - \sum_{i=1}^n \mu_i(\bsbb^{(j)}) (\bsbx_i^T\bsbb^{(j+1)}-\bsbx_i^T\bsbb^{(j)})
\\
=\frac{1}{2} (\bsbb^{(j+1)}-\bsbb^{(j)})^T \bsbmI(\bsbxi^{(j)}) (\bsbb^{(j+1)}-\bsbb^{(j)})
\end{eqnarray*}
%\vspace{-.4in}
%
%\noindent
for some $\bsbxi^{(j)}=\vartheta\bsbb^{(j)}+(1-\vartheta) \bsbb^{(j+1)}$ with $\vartheta \in (0, 1)$.
Therefore,
%\vspace{-.15in}
\begin{eqnarray*}
&&F(\bsbb^{(j+1)})+\frac{1}{2}(\bsbb^{(j+1)}-\bsbb^{(j)})^T(\bsbI- \bsbmI(\bsbxi^{(j)}))(\bsbb^{(j+1)}-\bsbb^{(j)}) \\
&=&  G(\bsbb^{(j)},\bsbb^{(j+1)}) \leq G(\bsbb^{(j)},\bsbb^{(j)})-\frac{C_1'}{2}(\bsbb^{(j+1)}-\bsbb^{(j)})^T  (\bsbb^{(j+1)}-\bsbb^{(j)})\\
&=&F(\bsbb^{(j)})-\frac{C_1'}{2}(\bsbb^{(j+1)}-\bsbb^{(j)})^T  (\bsbb^{(j+1)}-\bsbb^{(j)}).
\end{eqnarray*}
%\vspace{-.45in}
%
%\noindent
\eqref{asympreg-gen} follows from the following inequality
%\vspace{-.05in}
\begin{align*}
 F(\bsbb^{(j)})-F(\bsbb^{(j+1)}) \geq \frac{1}{2}(\bsbb^{(j+1)}-\bsbb^{(j)})^T \left(C_1' \bsbI + \bsbI - \bsbmI(\bsbxi^{(j)})\right)(\bsbb^{(j+1)}-\bsbb^{(j)}).
\end{align*}
%\vspace{-.4in}
%
%\noindent
%which proves \eqref{asympreg-gen}.

Now assume a subsequence $\bsbb^{(j_l)}\rightarrow\bsbb^*$ as $l\rightarrow\infty$. Under the condition $\rho< \max(1, 2-\max_k {\mathcal L}_{\Theta_k})$, $C>0$ and
\begin{align*}
\| \bsbb^{(j_l+1)}-\bsbb^{(j_l)}\|_2^2 \leq   (F(\bsbb^{(j_l)})- F(\bsbb^{(j_l+1)}))/C \leq   (F(\bsbb^{(j_l)})- F(\bsbb^{(j_{l+1})}))/C\rightarrow 0.
\end{align*}
That is, $\vec\Theta_k(\bsbb_k^{(j_l)}+\bsbX_k^T \bsby -\bsbX_k^T \bsbmu(\bsbb^{(j_l)});\lambda_k)-\bsbb_k^{(j_l)}\rightarrow 0$. From the continuity assumption,   $\bsbb^*$ is a group $\Theta$-estimate satisfying \eqref{thetaestglm-gen}.
 \qed

\bibliographystyle{elsarticle-harv}
\bibliography{glmbib}
%\end{spacing}
\end{document}